\documentclass{article}

\usepackage[nonatbib,preprint]{neurips_2026}
\usepackage[round]{natbib}

\usepackage[utf8]{inputenc} 
\usepackage[T1]{fontenc}    
\usepackage[table]{xcolor}         
\definecolor{spaceindigo}{HTML}{141B41}
\definecolor{duskblue}{HTML}{2A4F86}
\definecolor{steelblue}{HTML}{3F82CA}
\definecolor{cornflowerblue}{HTML}{6F9CEB}
\definecolor{babyblueice}{HTML}{98B9F2}
\definecolor{aliceblue}{HTML}{DFEAF6}
\definecolor{periwinkle}{HTML}{B6B5F7}
\definecolor{softperiwinkle}{HTML}{918EF4}
\definecolor{steelpurple}{HTML}{7B5FC8}
\definecolor{darkred}{RGB}{153, 0, 0}
\usepackage[pagebackref,breaklinks,colorlinks]{hyperref}
\hypersetup{
  colorlinks,
  linkcolor=darkred,     
  citecolor=steelblue,    
  urlcolor=steelblue     
}
\usepackage{url}            
\newcommand{\textcite}{\citet}
\newcommand{\parencite}{\citep}
\usepackage{caption}
\usepackage{booktabs}       
\usepackage{amsfonts}       
\usepackage{nicefrac}       
\usepackage{microtype}      
\usepackage{wrapfig}
\usepackage{enumitem}
\usepackage{amsthm}
\usepackage{amssymb}
\usepackage{mathtools}
\usepackage{multirow}
\usepackage{minitoc}
\usepackage{graphicx}
\usepackage[toc,page,header]{appendix}
\usepackage{tikz}
\usetikzlibrary{bayesnet, calc, arrows.meta, bending}
\usetikzlibrary{decorations.pathreplacing,angles,quotes,shapes.multipart, calligraphy}

\newcommand{\changelinkcolor}[1]{\hypersetup{linkcolor=#1}}
\AtBeginDocument{%
  
}

\theoremstyle{plain}

\theoremstyle{definition}

\theoremstyle{remark}

\DeclareMathOperator*{\argmax}{arg\,max}

\title{
Scaling Generative Foundation Models for\\Chest Radiography with Rectified Flow Transformers
}

\author{
Fabio De Sousa Ribeiro$^{1,2,\dagger,}$\thanks{Joint first authors{\enspace}$^\dagger$f.de-sousa-ribeiro@imperial.ac.uk} {\enspace} Emma A.M. Stanley$^{1,*}${\enspace}Charles Jones$^1${\enspace}Tian Xia$^1$
\\[3pt]
\textbf{Dominic C. Marshall$^{1,4}${\enspace}Laurent Renard Triché$^5${\enspace}Christopher V. Cosgriff$^{6,7}$}
\\[3pt]
\textbf{Panagiotis Dimitrakopoulos$^{2,3}${\enspace}Sotirios A. Tsaftaris$^{2,3}${\enspace}Ben Glocker$^{1,2}$}
\\[6pt]
$^1$Imperial College London,
$^2$Causality in Healthcare AI Hub, $^3$University of Edinburgh \\
$^4$Cleveland Clinic London,
$^5$Department of Perioperative Medicine, CHU Clermont-Ferrand
\\
$^6$Department of Medicine, Massachusetts General Hospital, $^7$Broad Institute of MIT and Harvard
\\[4pt]
Project page: \href{https://RadiT-project.github.io}{\texttt{RadiT-project.github.io}}
}

\begin{document}
\doparttoc
\faketableofcontents

\maketitle

\begin{abstract}
  We introduce the first generative foundation model for chest radiograph synthesis trained from scratch at the billion-parameter scale. Existing radiographic AI models often suffer from poor generalisation across patient subpopulations, institutions, and acquisition settings, resulting in limited real-world clinical utility. Controlled, high-fidelity synthesis of chest radiographs is a promising path toward diversifying clinical datasets and evaluating the robustness of diagnostic models. Therefore, we present the largest specialist generative foundation model for chest radiographs to date, with over 1.3B parameters, trained for 1.6T tokens on a curated, heterogeneous dataset comprising 1.2M radiographs and clinical expert-guided metadata.
  Our model supports controllable radiograph generation and editing across multiple demographic subgroups, acquisition views, and a dozen pathologies. Moreover, we significantly advance the state of the art in radiograph synthesis fidelity, producing images that are indistinguishable from real radiographs to clinical experts.
\end{abstract}

\section{Introduction}
\label{sec:introduction}
Real and diverse clinical datasets are difficult to collect and share at scale. Expert annotation is costly, pathology prevalence is highly skewed, and acquisition protocols vary substantially between institutions \parencite{topol_high-performance_2019,aristidou_bridging_2022}. Combined with privacy constraints, these factors impede the dissemination of medical data and hinder progress in healthcare AI \parencite{rieke2020future}. AI models continue to advance rapidly, yet they're often criticised for relying too heavily on statistical associations rather than capturing the underlying causal structure of the data \parencite{scholkopf2021toward,bareinboim2022pearl}. This limitation is particularly consequential in healthcare, as models that exploit spurious shortcuts generalise poorly across patient subpopulations, institutions, and acquisition settings \parencite{castro2020causality,d2022underspecification,jones_causal_2024}, ultimately undermining the utility of these systems in real-world clinical practice \parencite{cross2024bias,wenderott_facilitators_2025}. 

Alongside calls to expand and diversify clinical datasets~\parencite{seyyed2020chexclusion,arora2023value,hasanzadeh2025bias}, \textit{generative modelling} has recently emerged as a promising approach to addressing the aforementioned challenges~\parencite{ribeiro2023high,pezoulas2024synthetic,khosravi2024synthetically} by producing targeted synthetic examples that expand coverage of underrepresented subgroups~\parencite{moroianu2025improving}, facilitate model stress-testing~\parencite{perez2024radedit}, and improve robustness~\parencite{ROSCHEWITZ2025103668} and fairness in downstream tasks~\parencite{ktena2024generative}.

Despite longstanding speculation that AI would replace radiologists, radiology faces growing service pressures \parencite{afshari2025growing}, and chest radiography (CXR) remains the most commonly performed imaging examination worldwide. Recent advances in generative modelling have enabled realistic CXR synthesis from text reports \parencite{weber2023cascaded,perez2024radedit,bluethgen2025vision,moroianu2025improving}, alongside efforts to incorporate medical causal knowledge in principled ways \parencite{ribeiro2023high,ribeiro2025counterfactual}. However, existing CXR generative models are still limited in terms of fidelity \parencite{dutt2025chexgenbench}, demographic controllability, and causal consistency.
\textit{RoentGen-v2} \parencite{moroianu2025improving} begins to tackle these limitations, but is still constrained by its single-source training data from MIMIC-CXR \parencite{johnson2019mimic}, its restriction to single-view Posterior-Anterior (PA) radiographs, and its reliance on an adapted general-purpose Stable Diffusion v2.1 \parencite{rombach2022high} backbone, all of which hamper high-fidelity generation at scale. Although causal approaches to CXR synthesis exhibit impressive controllability \parencite{ribeiro2023high,xia2025decoupled,ribeiro2025counterfactual}, they still fall short of modern generative fidelity standards and have only been shown in lower-resolution, small-scale settings. Furthermore, generative models like \textcite{ribeiro2023high} rely on external classifiers for strong performance, risking the reintroduction of the very classifier bias they are intended to address \parencite{kumar2025prism}. Given these limitations, together with the potential of generative modelling to mitigate robustness and safety issues in healthcare AI \parencite{ma2025ai}, there is a clear need for more capable and scalable generative models for CXR.

\textbf{Contributions.} \; We present a generative foundation model for chest radiography at unprecedented scale and fidelity. 
To our knowledge, this is the first specialist generative model in chest radiography trained from scratch at the billion-parameter scale. 
Beyond advancing the state-of-the-art in CXR synthesis, our approach supports controllable generation across multiple demographic subgroups (e.g. age, race, sex), acquisition views (AP, PA, Lateral) and a dozen pathologies. 
Our contributions are:
\begin{enumerate}[label=(\roman*)]
    \item We collate the CXR7-1M dataset (\S\ref{sec:dataset}), the largest-scale open-source chest X-ray dataset to date, comprising over 1.2M radiographs, harmonised from multiple existing datasets and paired with radiologist-guided metadata systematically extracted through expert consultation.
    \item We introduce an expert‑designed, clinically plausible causal graph for CXR and instantiate it as the largest continuous flow-based causal model to date (\S \ref{subsec:flow_scm}), spanning 19 demographic and radiological variables, and unlocking scalable exact abduction for discrete factors.
    \item We ablate and train a series of scaled rectified flow transformers for chest X-ray generation, up to 1.3B parameters. Our largest model, \textbf{RadiT XL}, attains four-fold FDD and ten-fold KDD improvements over prior state-of-the-art on the CheXGenBench benchmark (\S\ref{subsec:fidelity}).
    \item We evaluate our models on controllable image generation and editing, showing that RadiT achieves high-fidelity control over multiple demographic, acquisition-view, and clinical attributes (\S\ref{subsec:control_image_exp}), producing radiographs that experts find indistinguishable from real ones.
\end{enumerate}
\section{Related Work}
\label{sec:related_work}
\textbf{Diffusion \& Flow Matching.} \; Diffusion models were introduced as discrete-time Markov processes that gradually transform data into noise, with a neural network parameterising the reverse process \parencite{sohl2015deep,NEURIPS2020_4c5bcfec}. Follow-up work improved their performance \parencite{nichol2021improved}, displacing GANs \parencite{goodfellow2014generative,karras2020analyzing} as the dominant approach \parencite{dhariwal2021diffusion}. \textcite{song2021scorebased} then unified diffusion and score-based models \parencite{NEURIPS2019_3001ef25} from an SDE perspective.
EDM \parencite{karras2022elucidating,Karras_2024_CVPR} and DiT \parencite{peebles2023scalable} refined latent diffusion \parencite{rombach2022high}, improving architectures and training.
More recently, Flow Matching \parencite{lipman2023flow,liu2023flow,albergo2023building} simplified and generalised diffusion-style modelling by enabling simulation-free training of continuous transports between \textit{arbitrary} distributions via neural ODEs \parencite{chen2018neural}. Most relevant to our work are Rectified Flows (RF)~\parencite{liu2023flow}, which seek to learn straight transport paths, offering conceptual simplicity and strong few-step sampling. RF transformers 
now power frontier generative models at scale \parencite{esser2024scaling,labs2025flux}, including our own.

\textbf{Chest Radiography Synthesis.} \; Previous work on chest radiography synthesis spans early task-specific generative models \parencite{madani2018chest,salehinejad2018generalization} and more recent report-conditioned foundation models. \textit{RoentGen} \parencite{chambon2022roentgen} helped establish latent diffusion CXR synthesis, showing that radiology reports can be used to generate realistic CXRs and improve downstream performance with synthetic data augmentation. Afterwards, \textit{ViewXGen} \parencite{lee2024vision} explored view-specific CXR generation, whereas \textit{Cheff} \parencite{weber2023cascaded} scaled the latent diffusion approach through a cascaded pipeline to enable higher resolution outputs. \textit{LLM-CXR} \parencite{lee2024llmcxr} extended the vision-language direction by combining CXR understanding and generation in a single instruction-tuned model. In parallel, a line of work on counterfactual image editing showed significant progress in CXR synthesis controllability \parencite{ribeiro2023high,ribeiro2025counterfactual,xia2024mitigating,xia2025decoupled,kumar2025prism}, inspiring applications in contrastive learning \parencite{ROSCHEWITZ2025103668}, stress-testing \parencite{perez2024radedit,ma2025ai}, and segmentation \parencite{mehta2025cf}. Recent work using diffusion models has further shown that synthetic radiographs can improve the robustness, fairness, and generalisation of downstream models \parencite{ktena2024generative}. \textit{RoentGen-v2} \parencite{moroianu2025improving} focuses on finer-grained control over radiographic findings and demographics, while \textit{ChexGen} \parencite{doi:10.1056/AIoa2500799} adds spatial control through masks and bounding boxes. 
Although \textit{ChexGen} is also trained from scratch on multiple CXR datasets, its generative pipeline is relatively dated by modern standards, relying on an SD-v1.5 VAE, staged 256-to-512 resolution training, and discrete-time diffusion instead of continuous-time rectified flow \parencite{esser2024scaling,labs2025flux}. Despite their clinical potential, existing CXR synthesis methods remain limited in fidelity and controllability. These limitations partly reflect insufficient data and model scaling, as well as the difficulty of building performant generative pipelines for clinical data, which this work addresses.
\begin{figure*}
    \centering
    \includegraphics[trim=0 0 0 0,clip,width=.99\textwidth]{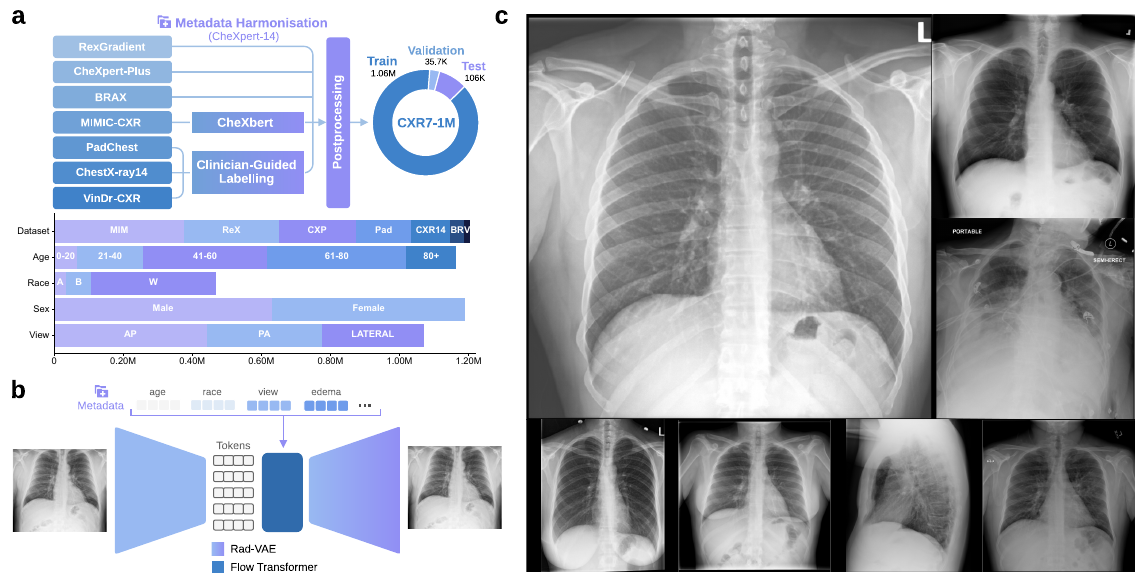}
    \caption{
    \textbf{Generative foundation model for chest radiography.}
    (\textbf{a}) Proposed CXR7-1M dataset, harmonised from seven existing datasets and augmented with additional radiologist-guided metadata.
    (\textbf{b}) Radiographic rectified flow transformer (RadiT), and VAE trained with Rad-DINO perceptual loss (Rad-VAE).
    (\textbf{c}) Synthetic 512${\times}$512 chest radiographs generated using our RadiT XL (1.3B).
    }
    \label{fig:figure1}
\end{figure*}
\section{CXR7-1M: Dataset Construction}
\label{sec:dataset}
The scale, diversity, and quality of training data are key determinants of AI capability \parencite{kaplan2020scaling,bommasani2021opportunities}. With this in mind, we devoted significant effort to the careful design, harmonisation, and expert‑in-the-loop curation of a large and diverse chest X‑ray dataset to improve generative synthesis. The resulting CXR7‑1M dataset comprises over 1.2 million radiographs paired with radiologist‑guided metadata (Figure \ref{fig:figure1}). To create CXR7-1M, we combine seven publicly available chest X-ray datasets, including MIMIC-CXR \parencite{johnson2019mimic}, CheXpert-Plus \parencite{irvin2019chexpert,chambon2024chexpert},  ReXGradient-160K \parencite{zhang2025rexgradient}, PadChest \parencite{bustos2020padchest}, VinDr-CXR \parencite{nguyen2022vindr}, NIH ChestX-ray14 \parencite{wang2017chestxray}, and BRAX \parencite{reis2022brax}. All available images were converted to 16-bit PNG with the aim of retaining radiometric fidelity while enabling efficient downstream use. Where available (i.e. MIMIC, CheXpert, VinDr, BRAX), we converted from original DICOM with correct windowing and photometric interpretation. For all images, pixel intensities were normalised to the full 16-bit range, before applying aspect-ratio preserving resampling, with centre padding where required, to obtain a consistent 512${\times}$512 image size. A detailed breakdown of the composition of CXR7-1M is provided in Appendix \ref{appsec:cxr7-1m}.
\subsection{Metadata Harmonisation with Clinical Guidance}
\label{subsubsec:metadata_harmonisation}
To enable controlled CXR synthesis, we performed a series of clinician-guided preprocessing and harmonisation steps to standardise pathology labels, patient demographics, and image acquisition metadata across all 7 datasets present in CXR7-1M. For consistent pathology labelling, we adopted the CheXpert-14 framework \parencite{irvin2019chexpert}, comprising 12 pathology categories in addition to `No Finding' and `Support Devices'. The pathology labels for different datasets were harmonised either by deriving them from radiology reports using CheXbert \parencite{smit2020combining}
or by manually mapping dataset-specific annotations to the 14 CheXpert labels in consultation with a clinical expert.

\textbf{Dataset-Specific Preprocessing.} \; For MIMIC-CXR, radiology report sections were processed using CheXbert to obtain CheXpert-14 compatible pathology labels. Patient `Age' and acquisition `View' information were extracted from the available metadata, and patient `Race' labels were consolidated to `Asian', `Black' and `White'. For PadChest, NIH, and VinDr-CXR, the original pathology labels were \textit{manually} mapped to the CheXpert-14 label space with the help of clinical experts, and any unmatched categories were marked as missing (`NaN'). For instance, PadChest contains 193 distinct pathology labels, each of which was individually mapped to an appropriate CheXpert-14 label and validated by a clinical expert. PadChest required `Age' information to be computed from study dates, and VinDr-CXR required majority voting across pathology annotations and extraction of `Age' and `Sex' from the original DICOM metadata. RexGradient already includes CheXbert-extracted labels, and `Age' was computed from study dates. CheXpert-Plus and BRAX were already broadly compatible and required minimal processing; missing/ambiguous values were assigned to `NaN'.

\textbf{Merged Metadata Filtering.} \; Following dataset-level harmonisation, we applied a set of global postprocessing steps to filter the merged metadata by: (i) replacing implausible ages (${>}110$) with `NaN'; (ii) restricting `Sex' to `Male'/`Female' labels; (iii) consolidating `Race' categories to `Asian', `Black' and `White'; (iv) standardising acquisition `View' positions to `AP', `PA' and `Lateral'; (v) treating all remaining invalid or ambiguous labels as missing and assigning them a `NaN' label. The resulting combined dataset was then split at the patient level into 302,871 training, 10,000 validation and 30,790 test subsets, corresponding to 1,063,255, 35,726, and 106,513 radiographs, respectively.

\section{Generative Foundation Model for Chest Radiography}
\label{sec:foundation}
We introduce a frontier generative foundation model for chest radiographs, designed for high-fidelity, controllable synthesis.
We return to first principles by designing and training domain-specific rectified flow models from scratch at the billion-parameter scale on the largest CXR dataset to date, thereby unlocking substantial improvements in fidelity. Next, we detail our model variants tailored to CXR.

\textbf{Preliminaries.} \; Flow Matching (FM) \parencite{lipman2023flow,liu2023flow,albergo2023building} provides a simulation-free way to train Continuous Normalizing Flows (CNFs) \parencite{chen2018neural}.
FM learns a map from a source $X_0 \sim p_{\text{src}}$ to a target distribution $X_1 \sim p_{\text{tgt}}$ via an ODE with neural velocity field $v : \mathbb{R}^d \times [0, 1] \to \mathbb{R}^d$. To train, we regress $v_t$ onto a known target velocity:
\vspace{-2pt} 
\begin{align}
     && \min_\theta \int_0^1 \mathbb{E}_{X_0 \sim p_{\text{src}}, X_1 \sim p_{\text{tgt}}} \left[\left\| v_t(X_t; \theta) - (X_1 - X_0)\right\|^2\right] \, \mathrm{d}t, && X_t = (1-t)X_0 + tX_1, &&
\end{align}
where $\mathrm{d}X_t = (X_1 - X_0) \, \mathrm{d}t$ defines the target velocity field of the prescribed straight-line interpolant. This simple choice of interpolant and objective is best known as a Rectified Flow \parencite{liu2023flow}. 
Samples are generated by drawing from the source distribution and integrating the learned ODE.
%
\subsection{Optimising Radiographic Perceptual Fidelity of VAEs}
\label{subsec:vae}
A key design component of large-scale latent flow/diffusion models is the VAE \parencite{esser2024scaling}. 
We systematically evaluate frontier VAEs for radiographic fidelity and, unlike prior work, begin by developing and training our own baseline VAE from scratch on CXR. We adopt a block design based on the EDM2 \parencite{karras2024analyzing} architecture, with magnitude-preserving layers for training stability. 
For complete architectural details and hyperparameters, please refer to Appendix \ref{appsec:rad_vae}.

\textbf{Radiographic Perceptual Training.} \; Our VAEs are trained with a domain-specific LPIPS perceptual loss inspired by \textcite{dino_perceptual}, which in their case replaces VGG with DINOv2/v3 features. However, we propose to use Rad-DINO \parencite{perez2025exploring} features instead to optimise for clinical, chest radiograph-specific perceptual quality. The VAE loss we optimise is:
\begin{align}
    \label{eq:perc}
    &&\mathcal{L}_{\text{VAE}} = \mathcal{L}_{\text{MSE}}(\mathbf{x}, \hat{\mathbf{x}}) + \beta D_{\text{KL}}(q_\phi(\mathbf{z} \mid \mathbf{x}) \parallel p(\mathbf{z})) + \alpha \mathcal{L}_{\text{Rad-LPIPS}}(\mathbf{x}, \hat{\mathbf{x}}), && \hat{\mathbf{x}} = D(E(\mathbf{x})), &&
\end{align}
where $E : \mathcal{X} \to \mathcal{Z}$ and $D : \mathcal{Z} \to \mathcal{X}$ denote the encoder and decoder networks respectively. We also benchmark several off-the-shelf VAEs pretrained on natural images (i.e. Stable Diffusion and FLUX), finding them to underperform our baseline VAE in terms of radiographic fidelity. 
To close this performance gap, we find that LoRA fine-tuning \parencite{hu2022lora} selected FLUX.2 VAE decoder layers using the proposed Rad-DINO perceptual training strategy in Eq. \eqref{eq:perc} performs best on CXR.

\subsection{Rectified Flow Architectures: Latent \& Pixel Spaces}
\label{subsec:flow_arch}
The design space we explore can be divided into two main parts: (i) whether to build a latent- or pixel-space flow model, with the former requiring a suitable VAE for CXR (see \S \ref{subsec:vae}); and (ii) the backbone architecture, i.e. U-Net or Transformer. Next, we present our model architecture variants.

\textbf{Vision Transformer \& U-Net Backbones.} \; The vision transformer architecture we use is built on DiT/SiT \parencite{peebles2023scalable,ma2024sit} with two upgrades to improve training stability and performance at scale. Namely, we add query-key RMS-Norm \parencite{zhang2019root} to the attention layers following \textcite{esser2024scaling}, and replace the standard MLPs with SwiGLU blocks \parencite{shazeer2020glu}. 
We use the same vision transformer architecture for both pixel- and latent-space variants of our flow models, with a patch size of 16 for the former and 2 for the latter.
Our largest flow transformer has over 1.3B parameters and is trained in the latent space of the FLUX.2 VAE.
For our U-Net model variants, we adopt the EDM2 backbone \parencite{karras2024analyzing} without modification, except for extending the channel multiplier schedule from 4 to 6 levels to support higher-resolution pixel-space modelling. For further architecture and training details, please refer to Appendix \ref{appsec:rect_flow}.

\textbf{Pixel-Space Flows.} \; 
Contrary to prior work on CXR synthesis, which exclusively builds on popular \textit{latent} diffusion pipelines, we find that pixel-space modelling can improve identity preservation in controlled generation settings, such as (counterfactual) image editing. 
To improve pixel-space training at high resolutions \parencite{esser2024scaling}, we use the shifted timestep mapping $t \mapsto t /(\alpha - t(\alpha -1))$ with $\alpha=3$, which is a time-inverted version of \textcite{labs2025flux}'s Logit-Normal parameterisation.

\textbf{Metadata Conditioning Strategy.} \; To enable controlled CXR synthesis, we need to effectively condition our generative model on all 19 metadata variables available in CXR7-1M. These include the CheXpert-14 pathology labels, the three patient demographic factors `Age', `Race',  and `Sex', acquisition view (with values `AP', `PA', `Lateral'), and a dataset source indicator (7 datasets), cf. \S\ref{sec:dataset}. 
All variables except for `Age' are categorical and are encoded with learned embedding tables. For each variable, class 0 is reserved as a null category for missing/ambiguous labels (`NaN'). We treat `Age' separately as a continuous variable, rescaling it by 100 and encoding it with the same Fourier/sinusoidal parameterisation used for timestep embeddings in the flow backbone. The `Age' encoding is then projected by an MLP to the shared embedding dimension $d$. The final conditioning embedding $\mathbf{z} \in \mathbb{R}^d$ is a magnitude-preserving scaled sum of the $n$ individual variable embeddings:
\begin{align}
    && \mathbf{z} = \frac{1}{\sqrt{n}} \sum_{i=1}^n \mathbf{e}^{(i)}, && \text{where} &&\mathbf{e}^{(i)} \coloneqq E\big(x^{(i)}; \phi^{(i)}\big) \in \mathbb{R}^d \quad \text{for all} \; i \in  \{1,\ldots,n\}, &&
\end{align}
with $E(\cdot, \phi^{(i)}) : \mathcal{X}^{(i)} \to \mathbb{R}^d$ denoting the embedding function for the $i^{\text{th}}$ variable $X^{(i)}$, parameterised by $\phi^{(i)}$. This prevents $\mathbf{z}$'s magnitude from growing excessively with $n$ and affecting training stability.
\subsection{Metadata Causal Modelling with Neural ODEs}
\label{subsec:flow_scm}
\begin{table}[!ht]
    \centering
    \sffamily
    \scriptsize
    \caption{
    \textbf{Clinical expert-informed causal graph of radiologic findings and demographic factors.} The causal parents of each finding are shown, with Age ({\sffamily \footnotesize A}), Race ({\sffamily \footnotesize R}), and Sex ({\sffamily \footnotesize S}) demographics.
    }
    \label{tab:causal_graph}
    \begin{tabular}{cccccccccccccc}
         \rotatebox{90}{\textcolor{steelblue}{\textbf{1}.} Atelectasis} & \rotatebox{90}{\textcolor{steelblue}{\textbf{2}.} Cardiomegaly} & \rotatebox{90}{\textcolor{steelblue}{\textbf{3}}. Consolidation} & \rotatebox{90}{\textcolor{steelblue}{\textbf{4}.} Edema} & \rotatebox{90}{\textcolor{steelblue}{\textbf{5}.} Enlarged CM} & \rotatebox{90}{\textcolor{steelblue}{\textcolor{steelblue}{\textbf{6}.}} Fracture} & \rotatebox{90}{\textcolor{steelblue}{\textbf{7}.} Lung Lesion} & \rotatebox{90}{\textcolor{steelblue}{\textbf{8}.} Lung Opacity} & \rotatebox{90}{\textcolor{steelblue}{\textbf{9}.} No Finding} & \rotatebox{90}{\textcolor{steelblue}{\textbf{10}.} Pleural Eff.} & \rotatebox{90}{\textcolor{steelblue}{\textbf{11}.} Pleural Other} & \rotatebox{90}{\textcolor{steelblue}{\textbf{12}.} Pneumonia} & \rotatebox{90}{\textcolor{steelblue}{\textbf{13}.} Pneumothorax} & \rotatebox{90}{\textcolor{steelblue}{\textbf{14}.} Support Dev.}
         \\
        \midrule
         \scriptsize\multirow{1}{*}{10,12} & \scriptsize\multirow{1}{*}{A,R,S} & \scriptsize\multirow{1}{*}{12} & \scriptsize\multirow{1}{*}{A,R,S} & \scriptsize\multirow{1}{*}{2} & \scriptsize\multirow{1}{*}{A,S} & \scriptsize\multirow{1}{*}{8} & \scriptsize \multirow{1}{*}{1,3,4,11} & \scriptsize\multirow{1}{*}{1-14} & \scriptsize \multirow{1}{*}{2,11,12} & \multirow{1}{*}{A,S} & \scriptsize\multirow{1}{*}{A,S} & \scriptsize\multirow{1}{*}{A,S,6} & \scriptsize\multirow{1}{*}{10,13} \\
    \end{tabular}
\end{table}
\textbf{Clinical Expert-Informed Causal Graph.} \; To incorporate clinical domain knowledge into the generative process of our CXR synthesis model, we developed a causal graph over the CXR7-1M metadata through iterative discussions with three experienced pulmonologists. As shown in Table~\ref{tab:causal_graph}, our causal graph includes `Age', `Race' and `Sex' demographic factors as well as all CheXpert-14 pathologies. 
For an in-depth clinical rationale for the selected edges, please refer to Appendix \ref{appsec:clinical_rationale}.

\textbf{Continuous-Time Flow SCM.} \; 
We build a Structural Causal Model (SCM) \parencite{pearl2009causality} of CXR7-1M metadata using the clinical expert-guided causal graph in Table \ref{tab:causal_graph}, such that CXR synthesis can reflect known clinical dependencies between demographic factors and pathologies.
An SCM specifies the data-generating process of $n$ variables $X_{1},X_{2}\ldots,X_{n}$ via deterministic functions: $X_{i} \coloneqq f_{i}(\mathbf{PA}_{i}, U_{i})$, where $\mathbf{PA}_{i}$ are $X_{i}$'s parents, and $U_{i}$ its exogenous noise variable. 
\textcite{pawlowski2020deep,ribeiro2023high} used classical Normalizing Flows to parameterise SCMs, but their approaches do not support deterministic abduction $U_{i} = f_{i}^{-1}(\mathbf{PA}_{i}, X_{i})$ for discrete variables.
To address this, we model each variable with a \textit{continuous-time} flow over $t\in[0,1]$ \parencite{ribeiro2025counterfactual}, but with all categorical variables $X_{i} \in \{1\ldots,K\}$ in one-hot space $Y_{i} = \operatorname{onehot}(X_{i}) \in \{0,1\}^K$: 
\begin{align}
    &&\mathrm{d}Y_{i}(t) = v_i(t,Y_i(t); \mathbf{PA}_{i})\,\mathrm{d}t, && Y_i{(0)} = U_{i}, && Y_i \coloneqq Y_i(1), && \text{for}~i=1,\dots,n.
\end{align}
We then train $n$ such flows jointly end-to-end with the following rectified flow matching objective:
\begin{align}
    \label{eq:fm_scm}
    \mathcal{L}_{\text{FM-SCM}}\coloneqq \frac{1}{n}\sum_{i=1}^n\int_0^1 \mathbb{E} \Big[\big\| v_{i}(t, Y_i(t); \mathbf{PA}_i) - (Y_i - U_i)\big\|^2\Big] \, \mathrm{d}t.
\end{align}
Samples are drawn ancestrally in a topological ordering of the causal graph, with continuous ODE solver outputs $\hat{Y}_{i} =U_i + \int_0^1 v_i(t, Y_i{(t)}; \mathbf{PA}_i) \mathrm{d}t \in \mathbb{R}^K$ being converted to discrete predictions $\hat{X}_{i}$ via an argmax over categories. For further details, please see Appendix \ref{appsec:metadata_scm}. \textcite{ribeiro2025counterfactual} provide the theoretical basis for causal identification in flow matching models that we rely on here.
\section{Experiments}
\label{sec:experiments}
Our experiments are organised into three main stages. First, to identify the most suitable VAE for CXR, we study the effect of the proposed radiographic perceptual loss. Second, we evaluate the fidelity of our generative models, comparing them with the state of the art and conducting an expert real-vs-synthetic discrimination study with three experienced pulmonologists.
Finally, we assess controllable editing capabilities of our models. To do this, we build and compare different patient metadata and clinical finding predictors for reliable subgroup identification. We then evaluate our flow models on effectively editing and preserving the identity of both real and sampled radiographs.

\textbf{Metrics.} \; We report FID/FDD \parencite{heusel2017gans} and KID/KDD \parencite{binkowski2018demystifying} metrics, as well as Precision, Recall, Density, and Coverage \parencite{naeem2020reliable} for generated image samples using Rad-DINO, DINOv3, and Inveptionv3. We use these same models to measure LPIPS perceptual distance between images.
For reconstruction quality, we use Structural Similarity Index Measure (SSIM), Peak Signal-to-Noise Ratio (PSNR), and Reconstruction Fréchet Distance (rFD).
For patient metadata prediction, we report Area Under the Receiver Operating Characteristic Curve (ROCAUC).
\subsection{Radiographic Perceptual Fidelity of VAEs}
\label{subsec:vae_exp}
\begin{wraptable}[12]{r}{0.62\linewidth}
    \vspace{-11pt}
    \footnotesize
    \centering
    \caption{
    \textbf{Performance comparison of radiographic VAEs.}
    Our RadVAEs are either trained from scratch or fine-tuned from a FLUX.2 base, using Rad-DINO perceptual loss (\S\ref{subsec:vae}).
    }
    \label{tab:vae_results}
    \vspace{-5pt}
    \resizebox{\linewidth}{!}{
    \begin{tabular}{lccccc}
        \toprule
        \multirow{2}{*}{\textbf{Model}} 
        & \multicolumn{3}{c}{\textbf{Reconstruction FD} $\downarrow$} 
        & \multirow{2}{*}{\textbf{PSNR} $\uparrow$} 
        & \multirow{2}{*}{\textbf{SSIM} $\uparrow$} \\
        & \scriptsize{Rad-DINO} 
        & \scriptsize{DINOv3} 
        & \scriptsize{Inceptionv3} 
        & & \\
        \midrule
        Stable Diffusion 2.1 
        & 0.3855 & 2.5369 & 0.6472 & 39.163 & 0.9542 \\
        Stable Diffusion XL 
        & 0.4475 & 1.7326 & 0.8849 & 39.886 & 0.9527 \\
        Stable Diffusion 3.5 
        & 0.0615 & 0.6820 & 0.3473 & 42.027 & 0.9329 \\
        FLUX.2 
        & 0.0887 & 0.6320 & 0.3848 & 45.413 & 0.9875 \\
        \midrule
        RadVAE (Scratch) 
        & \textbf{0.0476} & 2.2460 & 0.8938 & 43.706 & 0.9831 \\
        RadVAE{\scriptsize\textcolor{steelblue}{$^\text{FLUX.2}$}} (FT Head) 
        & 0.0616 & 0.6657 & 0.2358 & 45.555 & 0.9872 \\
        \rowcolor{babyblueice!20} 
        \textbf{RadVAE}{\scriptsize\textcolor{steelblue}{$^\text{FLUX.2}$}} (LoRA) 
        & 0.0487 & \textbf{0.5933} & \textbf{0.1879} & \textbf{45.829} & \textbf{0.9880} \\
        \bottomrule
    \end{tabular}
    }
\end{wraptable}
To identify the best VAE for CXR, we first train our baseline VAE from scratch on CXR7-1M, which we call RadVAE (Scratch).
For model architecture and training details, please see \S\ref{subsec:vae}, and Appendix \ref{appsec:experiments_radvae}.
We then benchmark state-of-the-art pretrained VAEs, such as Stable Diffusion 3.5 and FLUX.2 \parencite{esser2024scaling,labs2025flux}, and ablate VAE fine-tuning strategies using the proposed Rad-DINO perceptual loss in \S\ref{subsec:vae}. Since off-the-shelf VAEs were trained on RGB images and CXRs are greyscale, we repeat the input 3 times and average the decoder's RGB predictions. As reported in Tables \ref{tab:vae_results} \& \ref{apptab:vae_results}, RadVAE (Scratch) achieves the best radiographic fidelity (Rad-DINO rFD) while outperforming Stable Diffusion VAEs in terms of PSNR and SSIM. 
We find that averaging decoder RGB predictions achieves highly competitive results on CXR despite its simplicity when using the FLUX.2 VAE; observing an improved PSNR of 45.41 compared to our 43.71 RadVAE (Scratch) baseline.
To further improve FLUX.2 VAE performance on CXR, we systematically ablate VAE fine-tuning strategies using our Rad-DINO perceptual loss, which we denote by RadVAE{\scriptsize\textcolor{steelblue}{$^\text{FLUX.2}$}}. We find that LoRA fine-tuning \parencite{hu2022lora} the FLUX.2 VAE's \textit{Mid-Block}, and the final layer of each \textit{Up-Block}, combined with full fine-tuning of the RGB head, yields the best performance overall on CXR (cf. Tables \ref{tab:vae_results} \& \ref{apptab:vae_results}).
\begin{table}[!t]
    \footnotesize
    \centering
    \caption{
    \textbf{Comparative evaluation of CXR generative fidelity.} 
    All metrics were computed using Rad-DINO features.   
    Benchmark results are from \textit{CheXGenBench} \parencite{dutt2025chexgenbench}. We also report results on two internal test splits from \textit{CXR7-1M}, a MIMIC-CXR 5K split and a separate 50K split.
    Superscript ($^{\textcolor{steelblue}{\text{pix}}}$) denotes our flow model variants trained in pixel-space ($512{\times}512$ resolution). 
    }
    \label{tab:sample_results}
    \vspace{2pt}
    \begin{tabular}{lccccccr}
        \toprule
         \multirow{2}{*}{\textbf{Model}} & \textbf{FDD} $\downarrow$ & \textbf{KDD} $\downarrow$& \textbf{Precision} $\uparrow$ & \textbf{Recall} $\uparrow$ & \textbf{Density} $\uparrow$ & \textbf{Coverage} $\uparrow$ & \multirow{2}{*}{\textbf{Size}} \\
         & \tiny{(Rad-DINO)} & \tiny{(Rad-DINO)} & \tiny{(Rad-DINO)} & \tiny{(Rad-DINO)} & \tiny{(Rad-DINO)} & \tiny{(Rad-DINO)} &  \\
        \midrule
         SDv3.5 M & 91.302 & 0.103 & 0.632 & 0.205 &  0.401 & 0.244 & 2.5B \\
         LLM-CXR & 71.243 & 0.061 & 0.782 & 0.041 &  \textbf{0.671} & 0.459 & 12B \\
         RadEdit & 69.695 & 0.033 &  0.397 & 0.544 & 0.150 & 0.285 & 0.8B\\
         Pixart Sigma &  60.154 & 0.023 & 0.666 & 0.522 & 0.506 & 0.506 & 0.6B \\
         Sana & 54.225 & 0.016 & 0.674  & 0.614 & 0.520 & 0.548 & 0.6B \\
         \midrule
 RadiT B$^{\textcolor{steelblue}{\text{pix}}}$ & 29.259 & 0.0258 & 0.7449 & 0.3482
 & 0.5261
 & 0.4768
 & 0.3B \\
 RadiT B & 16.879 & 0.0050 & 0.7795 & 0.6732 
 & 0.6273
 & 0.5952
 & 0.3B \\
 RadUNet$^{\textcolor{steelblue}{\text{pix}}}$ & 25.417 & 0.0209 & 0.7459 & 0.4233
 & 0.5445
 & 0.5282
 & 0.3B \\
 RadUNet & 16.873 & 0.0051 & \textbf{0.7906} & 0.6818 
 & 0.6383
 & 0.6017
 & 0.5B \\
\textbf{RadiT XL} & \textbf{13.152} & \textbf{0.0013} & 0.7590 
  & \textbf{0.8602}
 & 0.5683
 & \textbf{0.6369}
 & 1.3B \\
 \midrule
\rowcolor{babyblueice!20}\hspace{-4pt}\textit{CXR7-1M (MIMIC)} & & & & & & & \\
 RadiT B$^{\textcolor{steelblue}{\text{pix}}}$ & 19.328 & 0.0125 & 0.9176 & 0.5314
 & 1.0970
 & 0.8373
 & 0.3B \\
     RadiT B & 12.200 & 0.0030 & 0.9144 & 0.7203
 & 1.1567
 & 0.9156
 & 0.3B \\
 RadUNet$^{\textcolor{steelblue}{\text{pix}}}$ & 16.976 & 0.0105 & 0.9141 & 0.5870
 & 1.0792
 & 0.8745
 & 0.3B \\
  RadUNet & 11.786 & 0.0027 & \textbf{0.9346} & 0.7388
 & \textbf{1.1967}
 & 0.9182
 & 0.5B \\
 \textbf{RadiT XL} & \textbf{9.128} & \textbf{0.0005} & 0.9056
  & \textbf{0.8532}
 & 1.0245
 & \textbf{0.9217}
 & 1.3B \\
\midrule
\rowcolor{babyblueice!20}\hspace{-4pt}\textit{CXR7-1M (50K)} & & & & & & & \\
    \textbf{RadiT XL} & 1.715 & 0.0005 & 0.8911
  & 0.8483
 & 0.9620
 & 0.8895
 & 1.3B \\
         \bottomrule
    \end{tabular}
\end{table}
\begin{figure*}
    \centering
    \includegraphics[width=.995\textwidth]{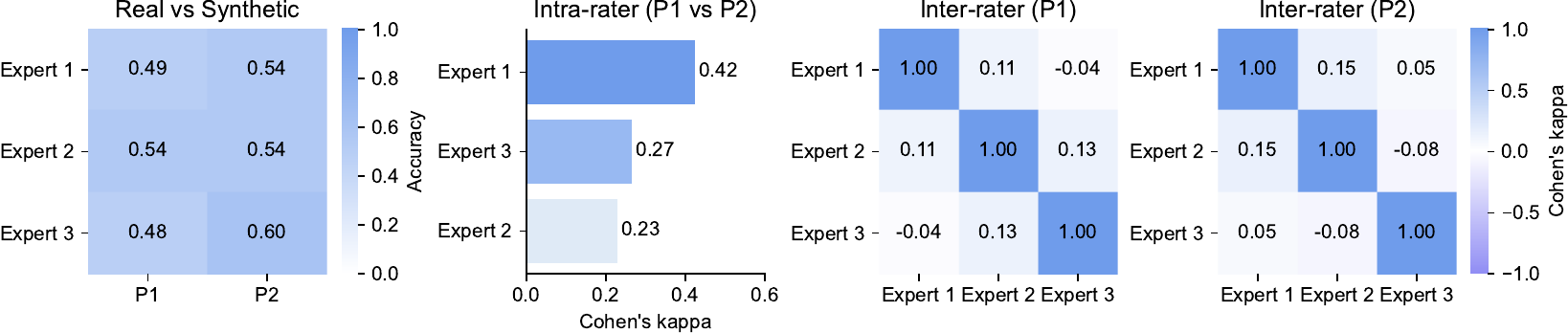}
    \caption{
    \textbf{Clinical experts' performance on the real-vs-synthetic task across 2 presentations.} 
    Near-chance accuracy and low intra- and inter-rater Cohen's $\kappa$ indicate high synthetic image realism.
    }
    \label{fig:ab_test}
\end{figure*}
\subsection{Evaluating Generative Fidelity: Benchmarks \& Clinical Expert Study}
\label{subsec:fidelity}
%
To evaluate the generative fidelity of our models, we first use CheXGenBench \parencite{dutt2025chexgenbench}.
The CheXGenBench test split has 5K samples from MIMIC-CXR, which we use as the closest available basis for fair comparison with prior work. We further evaluate our models on two internal test splits from the CXR7-1M dataset, a 5K MIMIC-CXR split and a separate 50K split. Tables \ref{tab:sample_results}, \ref{tab:sample_dinov3} and \ref{tab:sample_inception} report our results. Although direct comparison is limited by differences in training data (MIMIC-CXR vs CXR7-1M, see Table \ref{apptab:split_ablation}), our 1.3B flow transformer model (RadiT XL) achieves an FDD of 9.13, substantially improving on the previous best reported FDD of 54.23 under the available benchmark setting. This 83\% reduction in FDD signals a substantial gain in raw generative fidelity. In addition, RoentGen \parencite{chambon2022roentgen} and RoentGenv2 \parencite{moroianu2025improving} report FIDs of 96.1 and 76.8, respectively, on their MIMIC-CXR split, whereas our 1.3B model achieves an FID of 4.25 on ours, further indicating a large improvement in raw generative fidelity over recent CXR synthesis models. 
With a 50K reference test split from CXR7-1M, RadiT XL achieves an FDD of \textbf{1.72}, approaching the range of distributional distances typically seen in mature natural-image settings.

\textbf{Expert Real-vs-Synthetic Evaluation.} \; To further assess sample quality, we conducted a real-vs-synthetic discrimination experiment with three clinical experts.
We randomly selected 50 test-set images and generated 50 samples using our RadiT XL 1.3B model.
We measured accuracy against the ground truth, intra-rater consistency across 2 presentations, and pairwise inter-rater agreement using Cohen's $\kappa$. As reported in Figure \ref{fig:ab_test}, expert accuracy was near chance, with low intra- and inter-rater agreement, indicating that synthetic images were almost indistinguishable from real ones.
\begin{figure*}[!t]
    \centering
    \includegraphics[width=\textwidth]{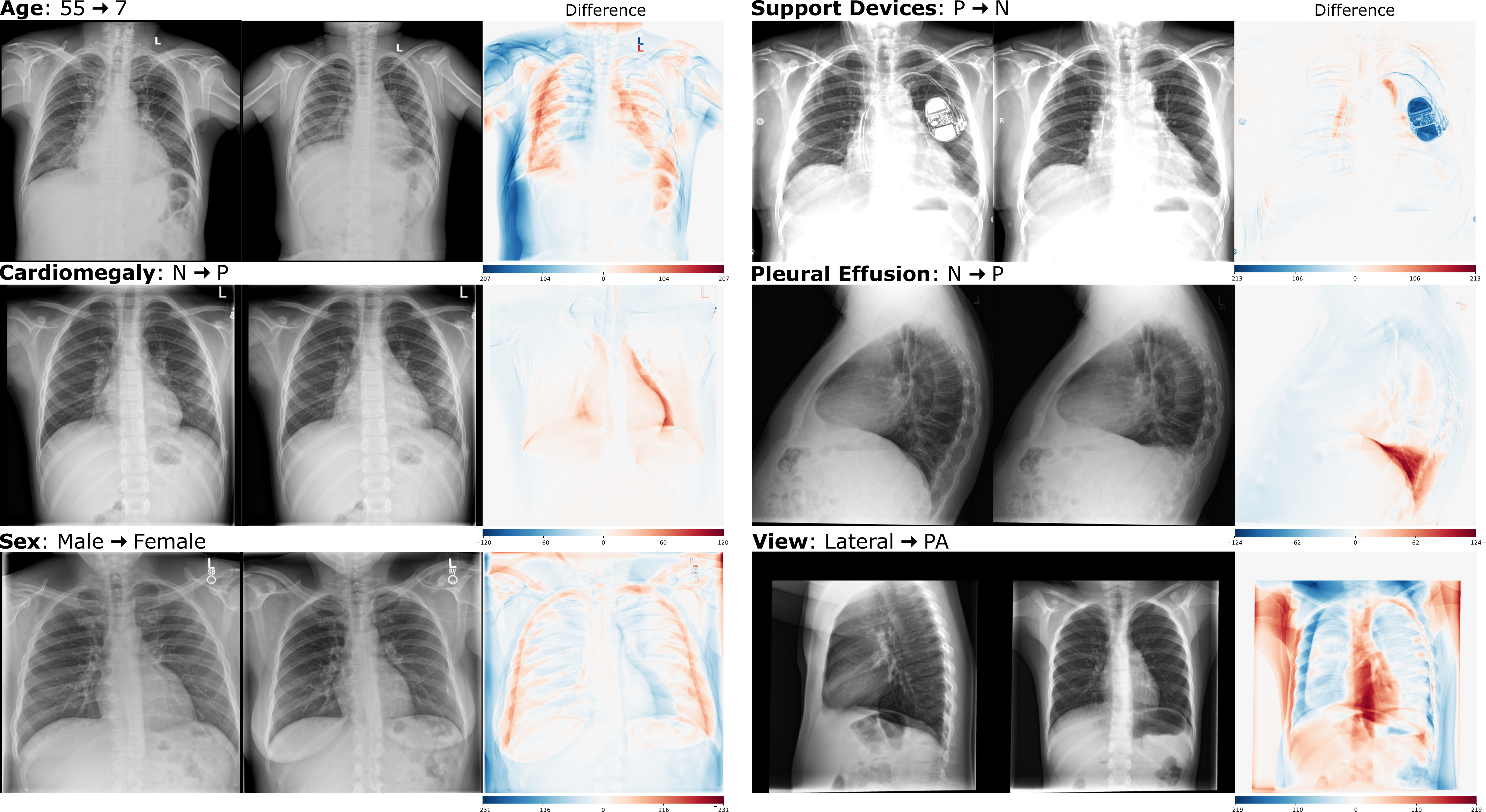}
    \caption{Selected 512${\times}$512 edits of real radiographs generated by our RadiT XL (1.3B) model.}
    \label{fig:quali_real_edits}
\end{figure*}
\subsection{Controllable Image Generation: Subgroup Analysis}
\label{subsec:control_image_exp}
\begin{wrapfigure}[16]{r}{0.5\textwidth}
    \vspace{-9pt}
    \centering
    \includegraphics[trim=0 0 0 0,clip,width=.48\textwidth]{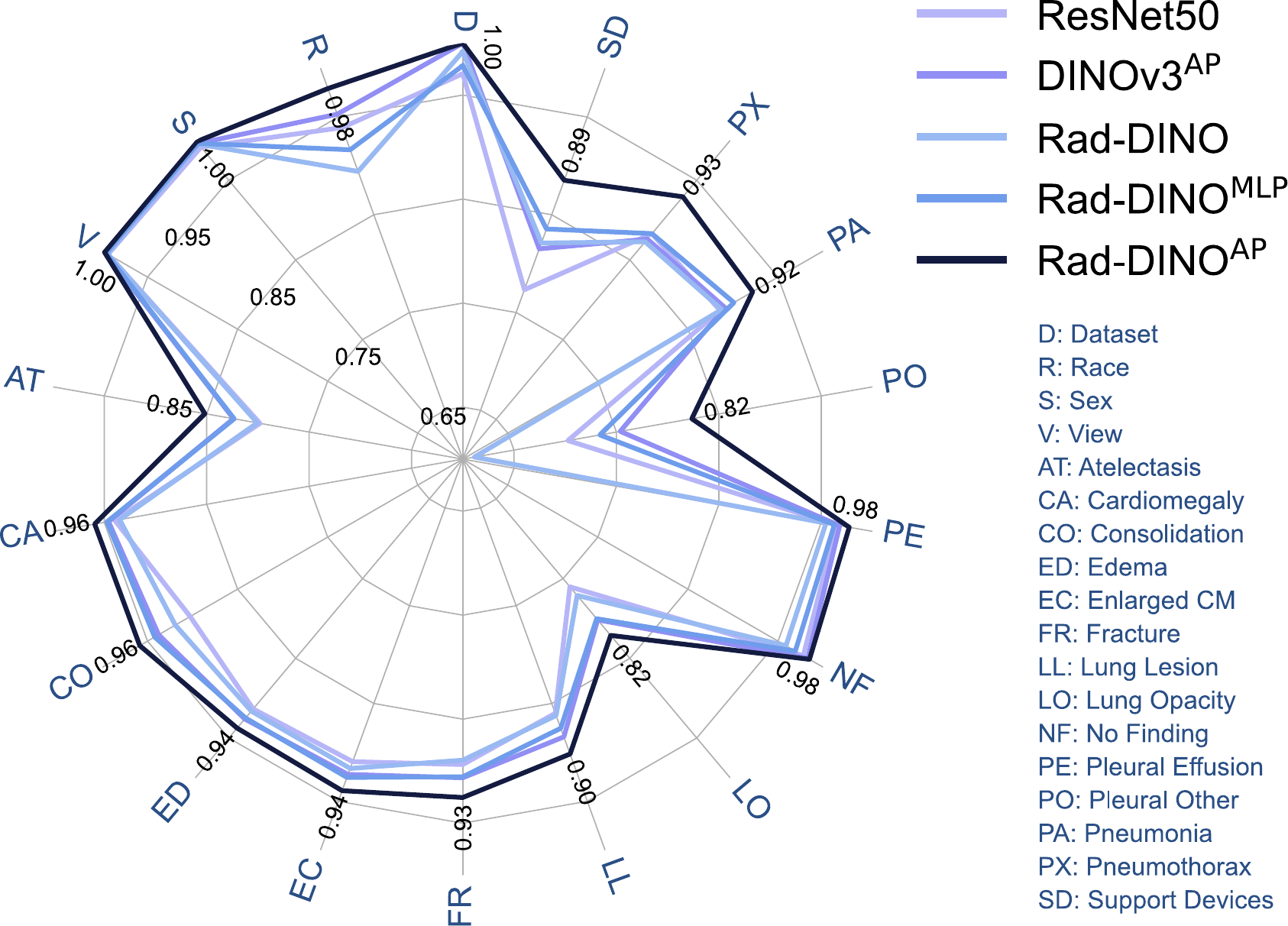}
    \caption{
        ROCAUC performance radar plot of our patient metadata predictors trained on CXR7-1M.
    }
    \label{fig:rad_dino_ap}
\end{wrapfigure}
\textbf{Patient Metadata Predictors.} \
To facilitate subgroup analysis of our models, we develop patient metadata predictors on CXR7-1M.
These predictors may be of independent interest to the community due to their strong performance.
We consider all 19 target variables in CXR7-1M (\S\ref{sec:dataset}), and train our predictors with a multi-task loss (Cross-Entropy/MSE) that automatically skips tasks with missing labels.
We first establish a baseline with a fully fine-tuned ImageNet-pretrained ResNet50. We then train three Rad-DINO models: (i) a linear probe on the 768-dimensional CLS embeddings; (ii) the same setup augmented with an MLP adapter; (iii) an attention pooling (AP) model over concatenated multi-layer patch features, inspired by \textcite{ilse2025data}. For comparison, we also train a DINOv3 model with the same AP setup. As shown in Figure~\ref{fig:rad_dino_ap}, Rad-DINO$^{\text{AP}}$ outperforms all baselines, particularly on patient `Race' classification.

\begin{figure*}
    \centering
    \includegraphics[trim=0 5 5 0,clip,width=\textwidth]{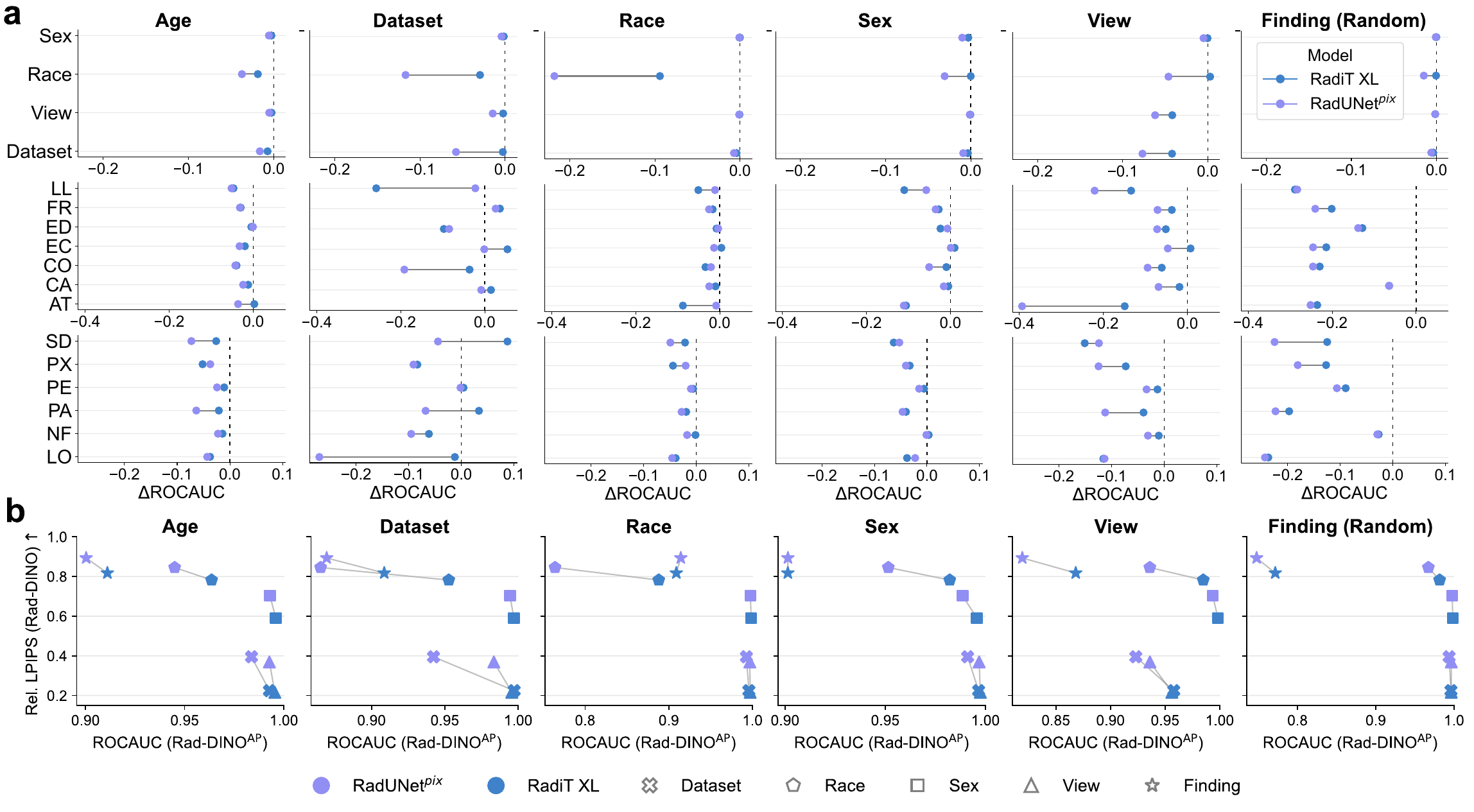}
    \caption{
    \textbf{Performance evaluation of subgroup-controlled CXR synthesis.}
    Plot titles denote edited attributes. (\textbf{a}) Difference in Rad-DINO$^{\text{AP}}$ ROCAUC between 5K CXR7-1M images and their edits.
    $\Delta$ROCAUC values closer to zero indicate better performance. (\textbf{b})
    Trade-off between effective editing (\textsc{x}-axis) and patient identity preservation (\textsc{y}-axis) for our best latent- vs. pixel-space flow models.
    }
    \label{fig:edit_auc}
    \hfill
\end{figure*}

\textbf{Controlled Editing.} \; We perform controlled editing of patient metadata and clinical findings for a random test subset of 5K patients from CXR7-1M, and use Rad-DINO$^{\text{AP}}$ to evaluate the ability of our generative models (RadUNet$^{\textcolor{steelblue}{\text{pix}}}$ and RadiT XL) to produce the desired edits. We intervene on each of the patient metadata attributes (sex, race, age, view, dataset), as well as on random finding categories from CheXpert-14, for six sets of experiments in total. To ensure the resulting counterfactual images are plausible and in-distribution for CXR7-1M, we use our Flow SCM (\S\ref{subsec:flow_scm}) to specify the full metadata and clinical finding profiles for a given intervention. As shown in Figure \ref{fig:edit_auc}, RadiT XL consistently outperforms RadUNet$^{\textcolor{steelblue}{\text{pix}}}$ on image editing.
We also perform targeted editing on \textit{random samples}, as this takes half the ODE solving time to generate image pairs.
To do this, we fix an exogenous noise vector representing a distinct identity, and perform edits using our Flow SCM (\S\ref{subsec:flow_scm}). We observe similar trends in editing performance compared to real images (cf. Figure \ref{appfig:sample_edit_auc}).
%

\textbf{Patient Identity Preservation.} \; 
 Patient identity preservation is the degree to which non-intervened upon image attributes remain unaffected by controlled edits. 
 We evaluate this by computing similarity between original and edited CXRs in pixel-space (SSIM), deep feature-space (LPIPS), as well as attribute-specialised latent space. The latter uses cosine similarity of metadata-tuned embeddings extracted from the individual task prediction heads of Rad-DINO$^{\text{AP}}$. 
 We observe a trade-off between effective editing and patient identity preservation in pixel- vs latent-space flow models (Fig.~\ref{fig:edit_auc},\ref{fig:cos_sim_bigmodels}). We find that RadUNet$^{\textcolor{steelblue}{\text{pix}}}$ outperforms RadiT XL in terms of identity preservation despite underperforming it in terms of generative fidelity and edit effectiveness (Table \ref{apptab:identity_pres}). To assess whether latent encoding is the major contributor to a loss of identity, we also compare approximately parameter-matched pixel and latent flow models for both the transformer and UNet backbone, and find that pixel-space flows generally outperform latent-space flows on our identity preservation metrics (Appendix \ref{appsec:experiments_control}).
\section{Conclusion}
\label{sec:conclusion}
We introduced the first specialist generative foundation model for chest radiography trained from scratch at the billion‑parameter scale, alongside CXR7‑1M, the largest open-source chest X‑ray dataset to date, curated with systematic clinical expert consultation.
Our largest model, \textbf{RadiT XL}, produces synthetic radiographs that were indistinguishable from real ones to a panel of clinical experts. In addition, it supports metadata-controlled synthesis and clinically coherent manipulation of findings informed by expert‑designed causal structure.
Limitations of this work include reliance on CheXpert-14 labels, which are unavailable for some CXR7-1M datasets, and incomplete negative-label coverage, as clinical reports often leave absent findings unstated rather than explicitly marking them as negative.
Additionally, while we propose novel approaches to measuring identity preservation in image editing, robustly measuring patient identity from radiographs alone remains challenging and may require clinical input alongside higher-quality metadata.
Our work also opens several directions for future research. First, large‑scale, high‑fidelity generative models can serve as testbeds for studying shortcut reliance, causal validity, and failure modes of medical imaging systems under controlled distribution shifts. Second, the combination of expert‑guided metadata and continuous‑time causal modelling provides a foundation for integrating medical knowledge into scalable generative pipelines, bridging symbolic (clinical) reasoning and modern foundation models. Finally, CXR7‑1M and RadiT XL create opportunities for downstream applications such as fairness auditing, representation learning, synthetic cohort construction, and cross‑institutional generalisation studies that are difficult to realise with real data alone.
We hope that releasing both the dataset and models will accelerate research on trustworthy, robust, and clinically grounded medical AI, and serve as a blueprint for scaling domain‑specific generative foundation models in other areas of healthcare imaging and beyond.

\begin{ack}
This project was supported by the UKRI AI programme, and the EPSRC, for CHAI - EPSRC Causality in Healthcare AI Hub [EP/Y028856/1], by the Royal Academy of Engineering as part of the Kheiron/RAEng Research Chair, by the Centre of Excellence for Regulatory Science \& Innovation in AI \& Digital Health Project Support Fund [ref PSF04], and through Imperial's UKRI Impact Acceleration Account [EP/X52556X/1]. DCM is supported by an MRC clinical research training fellowship [MR/Y000404/1] and the Mittal Fund at Cleveland Clinic Philanthropy. CVC is supported by an NIH grant [T32HL116275]. The authors acknowledge the use of resources provided by the Isambard-AI National AI Research Resource (AIRR). Isambard-AI is operated by the University of Bristol and is funded by the UK Government’s Department for Science, Innovation and Technology (DSIT) via UKRI; and the Science and Technology Facilities Council [ST/AIRR/I-A-I/1023].
\end{ack}

\bibliographystyle{plainnat}
\setlength{\bibsep}{.9\baselineskip}
\bibliography{bibfile}

\newpage
\appendix
\onecolumn
\addcontentsline{toc}{section}{Appendices}
\part{Appendices}
\changelinkcolor{black}{}
\parttoc

\section{CXR7-1M: Dataset Composition}
\label{appsec:cxr7-1m}
\begin{figure}[!ht]
    \centering
    \includegraphics[trim=5 0 5 5,clip,width=0.95\textwidth]{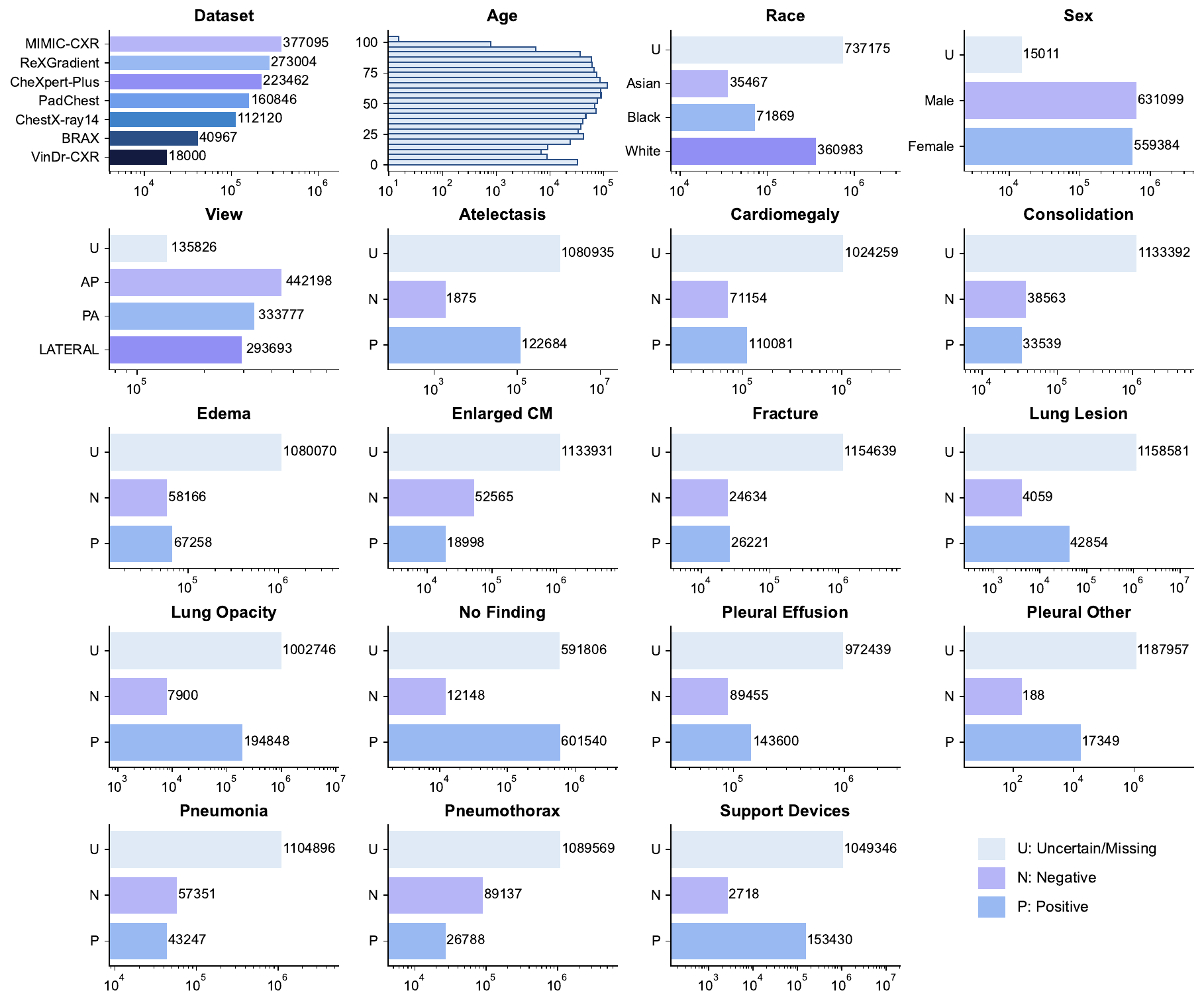}
    \caption{\textbf{CXR7-1M dataset composition.} Showing the 7 different dataset sources (top left) and all the available metadata variables, which were harmonised with the help of a clinical expert.}
    \label{appfig:cxr7_1m_stats}
\end{figure}
\section{Generative Foundation Model for Chest Radiographs}
\label{appsec:rad_foundation}
\subsection{Radiographic VAE Architecture}
\label{appsec:rad_vae}
\begin{figure*}[!ht]
    \centering
\includegraphics[width=.7\textwidth]{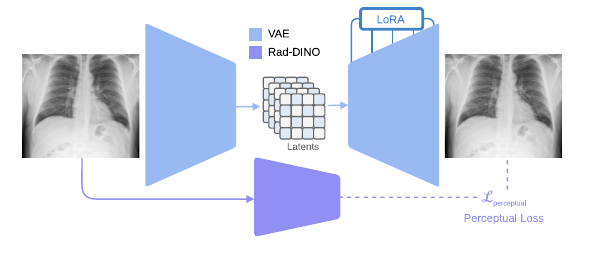}
    \caption{
    \textbf{RadVAE radiographic perceptual training schematic.} Our RadVAE variants are optimised for radiographic fidelity by either training from scratch or LoRA fine-tuning from a FLUX.2 base, using a domain-specific Rad-DINO perceptual loss (\S\ref{subsec:vae}).
    }
    \label{fig:vae}
\end{figure*}
\begin{table}[!ht]
    \small
    \centering
    \caption{Architecture hyperparameters for the proposed RadVAE, trained from scratch on CXR7-1M.}
    \vspace{2pt}
    \begin{tabular}{lr|lr}
        \toprule
        \multicolumn{4}{c}{\textbf{RadVAE Architecture}} \\[2pt]
        Setting & Value & Setting & Value \\
        \midrule
        Image resolution & 512${\times}$512 & Residual blocks (per stage) & 3 \\
        Image channels & 1 & Attention resolutions & - \\
        Latent resolution & 64 & Attention head channels & 64 \\
        Latent channels & 16 & Attention balance & 0.3 \\
         Model channels & 128 & Residual balance & 0.3 \\
        Channel multipliers & [1, 2, 3, 4] & Dropout & 0.0\\
        \midrule
        \#Parameters & \multicolumn{3}{r}{0.8B} \\
        \bottomrule
    \end{tabular}
    \label{apptab:rad_vae_arch}
\end{table}
\paragraph{RadVAE.} Table \ref{apptab:rad_vae_arch} reports the hyperparameters used to build RadVAE, which has 88M parameters. To help identify the best VAE to use for CXR, we first build and train our own VAE from scratch on CXR7-1M, which we call RadVAE. To that end, we adopt a block design based on the EDM2 \parencite{karras2024analyzing} architecture, with magnitude-preserving layers for stable training.
The main modifications we make are: (i) adding magnitude-preserving 1${\times}$1 convolutions after the encoder output to obtain bottleneck representations, and before the decoder input to project these representations back to the decoder feature space; (ii) removing all encoder-decoder skip-connections to maximise the information content of the bottleneck representation. It is also worth mentioning that the base EDM2 architecture applies attention only at the decoder bottleneck, whereas our variant applies bottleneck attention symmetrically in both the encoder and decoder. Following modern VAEs \parencite{labs2025flux}, we choose a wider latent bottleneck of 16 channels and keep the spatial resolution at 64${\times}$64.
\subsection{Flow Model Architectures for CXR}
\label{appsec:rect_flow}
\begin{figure*}[!ht]
    \centering
\includegraphics[width=.95\textwidth]{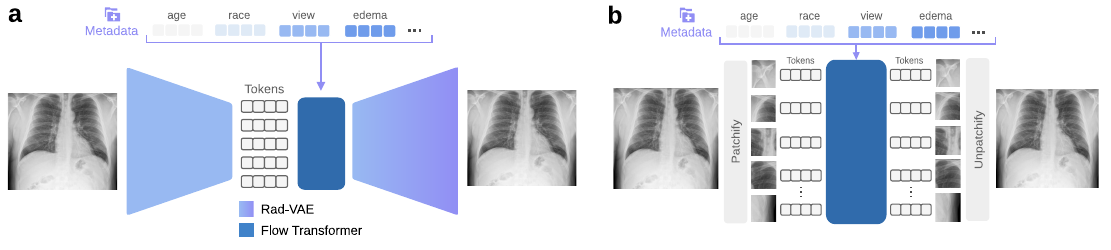}
    \caption{\textbf{Rectified flow transformer architectures.} 
(\textbf{a}) Latent-space rectified flow models operate on Rad-VAE latent tokens and use patient metadata conditioning to generate controllable chest radiographs.
(\textbf{b}) Pixel-space rectified flow models operate directly on image patch tokens with the same metadata conditioning interface, avoiding an explicit VAE bottleneck.
}
    \label{fig:models}
\end{figure*}
\begin{table}[!ht]
    \small
    \centering
    \caption{Transformer architecture hyperparameters for our CXR flow models (RadiT). 
    }
    \vspace{2pt}
    \begin{tabular}{l|ccc}
        \toprule
        \multirow{2}{*}{Setting} & \textbf{RadiT B}$^{\textcolor{steelblue}{\text{pix}}}$ & \textbf{RadiT B} & \textbf{RadiT XL} \\
        & (Pixel)  & (Latent) & (Latent) \\
        \midrule
        Input resolution & 512${\times}$512 & 64${\times}$64 & 64${\times}$64 \\
        Input channels & 1 & 32 & 32 \\
        Patch size & 16 & 2 & 2 \\
        Hidden size & 1024 & 1024 & 1536 \\
        Depth & 16 & 16 & 32 \\
        Num. heads & 16 & 16 & 24 \\
        Head dim. & 64 & 64 & 64  \\
        QK RMS-Norm & Yes & Yes & Yes  \\
        MLP block & SwiGLU & SwiGLU & SwiGLU \\
        MLP ratio & 8/3 & 8/3 & 8/3 \\
        Metadata conditions & 19 & 19 & 19 \\
        VAE Encoder & None & FLUX.2 & FLUX.2 \\
        VAE Decoder & None & RadVAE{\scriptsize\textcolor{steelblue}{$^\text{FLUX.2}$}} & RadVAE{\scriptsize\textcolor{steelblue}{$^\text{FLUX.2}$}} \\
        \midrule
        \#Parameters & 0.3B & 0.3B & 1.3B \\
        \bottomrule
    \end{tabular}
    \label{apptab:radit}
\end{table}
\textbf{Vision Transformer Backbone.} \; Table \ref{apptab:radit} provides the transformer architecture hyperparameters for our CXR flow models.
The vision transformer \parencite{dosovitskiy2021an} architectures we use are based on DiT/SiT \parencite{peebles2023scalable,ma2024sit} and include two modernising upgrades to improve training stability. Specifically, we add query-key RMS-Norm as suggested by \textcite{esser2024scaling}, and replace the standard MLP blocks with SwiGLU blocks \parencite{shazeer2020glu}. This choice is further supported by the adoption of SwiGLU in recent high-performing LLM architectures \parencite{grattafiori2024llama,liu2024deepseek}. We build three CXR flow transformer variants: (i) a 0.3B parameter pixel-space (512${\times}$512) flow transformer (RadiT B$^{\textcolor{steelblue}{\text{pix}}}$); (ii) the same as (i) but trained in the 32${\times}$64${\times}$64 latent space of a FLUX.2 VAE (RadiT B); (iii) the same as (ii) but much larger, with 1.3B parameters (RadiT XL). For reference, DiT XL/2 has 0.6B parameters \parencite{peebles2023scalable}.
\begin{table}[!ht]
    \small
    \centering
    \caption{UNet architecture hyperparameters for our CXR flow models (RadUNet). 
    }
    \vspace{2pt}
    \begin{tabular}{l|cc}
        \toprule
        \multirow{2}{*}{Setting} & \textbf{RadUNet}$^{\textcolor{steelblue}{\text{pix}}}$ & \textbf{RadUNet} \\
        & (Pixel)  & (Latent) \\
        \midrule
        Input resolution & 512${\times}$512 & 64${\times}$64  \\
        Input channels & 1 & 32 \\
        Model channels & 32 & 256 \\
        Channel multipliers & [1, 2, 4, 8, 16, 32] & [1, 2, 3 ,4] \\
        Residual blocks (per stage) & 3 & 3 \\
        Attention resolutions & [32, 16] & [16, 8] \\
        Attention head channels & 64 & 64 \\
        Dropout & 0.0 & 0.1 \\
        Metadata conditions & 19 & 19 \\
        VAE Encoder & None & FLUX.2 \\
        VAE Decoder & None & RadVAE{\scriptsize\textcolor{steelblue}{$^\text{FLUX.2}$}} \\
        \midrule
        \#Parameters & 0.3B & 0.5B \\
        \bottomrule
    \end{tabular}
    \label{apptab:radunet}
\end{table}
\paragraph{U-Net Backbone.} Table \ref{apptab:radunet} provides the transformer architecture hyperparameters for our CXR flow models.
As outlined in the main text (\S\ref{subsec:flow_arch}), the U-Net \parencite{ronneberger2015u} architectures we use are based on EDM2 \parencite{karras2024analyzing,karras2022elucidating}. Since these architectures are already highly optimised, we found that any further architectural modifications were unnecessary. We build two CXR flow U-Net variants: (i) a 0.3B parameter pixel-space (512${\times}$512) flow U-Net (RadUNet$^{\textcolor{steelblue}{\text{pix}}}$); (ii) an EDM2-M model \parencite{karras2024analyzing} trained in the 32${\times}$64${\times}$64 latent space of a FLUX.2 VAE (RadUNet). We note that RadUNet$^{\textcolor{steelblue}{\text{pix}}}$ departs from the architecture presets by \textcite{karras2024analyzing}, as the channel multiplier schedule is extended from 4 to 6 levels to support pixel-space modelling.
\subsection{Metadata Causal Model: Continuous-Time Flows}
\label{appsec:metadata_scm}
\begin{figure}[!ht]
    \centering
    \begin{tikzpicture}
    [
    scale=1, 
        every node/.style={
        scale=.7,
        rectangle,
        rounded corners=3pt,
        fill=steelblue!10,
        minimum width=2.5cm,
        minimum height=1cm,
        align=center,
        text=steelblue,
        font=\sffamily
    },
    every path/.style={
        draw=steelblue!55
    }
    ]
    
    \node[fill=periwinkle!30, text=steelpurple] (age) {\text{Age}};
    \node[right=49.4pt of age, fill=periwinkle!30, text=steelpurple] (race) {\text{Race}};
    \node[right=49.4pt of race, fill=periwinkle!30, text=steelpurple] (sex) {\text{Sex}};
    \node[below left=27pt and 15pt of age] (cardiomegaly) {\text{Cardiomegaly}};
    \node[right=15pt of cardiomegaly] (edema) {\text{Edema}};
    \node[right=15pt of edema] (pleural_other) {\text{Pleural Other}};
    \node[right=15pt of pleural_other] (pneumonia) {\text{Pneumonia}};
    \node[right=15pt of pneumonia] (fracture) {\text{Fracture}};
    \node[right=15pt of fracture] (pneumothorax) {\text{Pneumothorax}};

    \node[below=25pt of cardiomegaly] (enlarged_cm) {\text{Enlarged CM}};
    \node[right=50pt of enlarged_cm] (consolidation) {\text{Consolidation}};
    \node[below=25pt of pneumothorax] (support_devices) {\text{Support Devices}};
    \node[left=50pt of support_devices] (pleural_effusion) {\text{Pleural Effusion}};

    \node[below=109.75pt of sex] (atelectasis) {\text{Atelectasis}};
    \node[below=109.75pt of race] (lung_opacity) {\text{Lung Opacity}};
    \node[below=109.75pt of age] (lung_lesion) {\text{Lung Lesion}};

    \edge[-stealth]{race}{edema}
    \edge[-stealth]{race}{cardiomegaly}

    \edge[-stealth]{age}{cardiomegaly}
    \edge[-stealth]{age}{pleural_other}
    \edge[-stealth]{age}{edema}
    \edge[-stealth]{age}{pneumonia}
    \edge[-stealth]{age}{fracture}
    \edge[-stealth]{age}{pneumothorax}

    \edge[-stealth]{sex}{cardiomegaly}
    \edge[-stealth]{sex}{pleural_other}
    \edge[-stealth]{sex}{edema}
    \edge[-stealth]{sex}{pneumonia}
    \edge[-stealth]{sex}{fracture}
    \edge[-stealth]{sex}{pneumothorax}
    
    \edge[-stealth]{cardiomegaly}{enlarged_cm}
    \edge[-stealth]{cardiomegaly}{pleural_effusion}
    \edge[-stealth]{pleural_other}{pleural_effusion}
    \edge[-stealth]{pneumonia}{consolidation}
    \edge[-stealth]{pneumonia}{pleural_effusion}
    \edge[-stealth]{pleural_other}{lung_opacity}
    \edge[-stealth]{fracture}{pneumothorax}
    \edge[-stealth]{edema}{lung_opacity}  
    \edge[-stealth]{pleural_effusion}{support_devices}
    \edge[-stealth]{pleural_effusion}{atelectasis}
    \edge[-stealth]{pneumothorax}{support_devices} 
    \edge[-stealth]{pneumothorax}{atelectasis}

    \edge[-stealth]{lung_opacity}{lung_lesion}
    \edge[-stealth]{atelectasis}{lung_opacity}
    \edge[-stealth]{consolidation}{lung_opacity}

    \end{tikzpicture}
    \caption{Proposed clinical expert-informed causal graph of demographic factors and radiologic findings. The graph was developed through iterative discussions with three experienced pulmonologists.}
    \label{fig:causal_graph}
\end{figure}
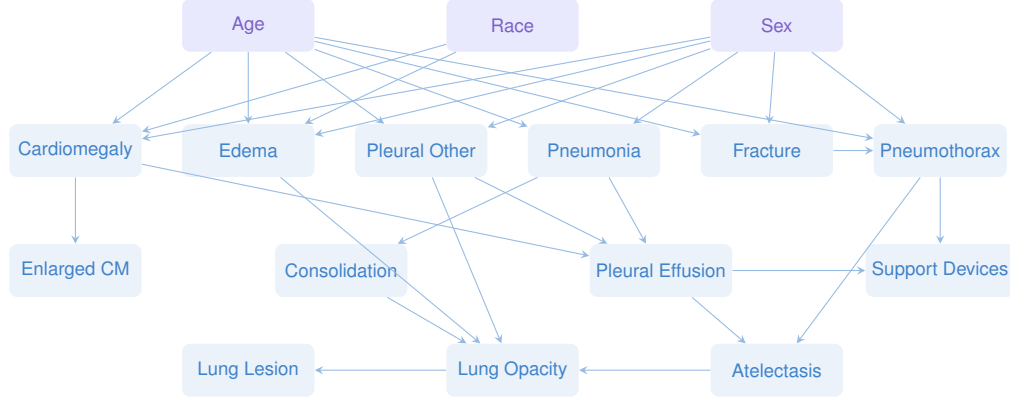

\subsubsection{Clinical Rationale for the Causal Graph}
\label{appsec:clinical_rationale}

Edge directionality was determined by two complementary principles. Three experienced pulmonologists reviewed the graph in an iterative process, reaching consensus on all included edges before finalisation. First, demographic factors were positioned as upstream roots: `Age', `Sex', and `Race' are known to modulate the prevalence of structural cardiac disease, pulmonary infection, osteoporotic fracture, and fluid retention, and therefore act as common causes of several downstream findings rather than consequences of them. 
Second, we encoded a clinical hierarchy among the findings themselves, reflecting the mechanistic sequence by which radiographic abnormalities arise. For example, pneumonia produces consolidation, which manifests as a lung opacity; a rib fracture can cause pneumothorax through pleural disruption; cardiomegaly, as a marker of underlying cardiac dysfunction, predisposes to pulmonary oedema and pleural effusion; and pleural effusion promotes atelectasis through compressive collapse. Lung lesion occupies the terminal node of the graph, representing a non-specific radiographic disturbance that may be the downstream expression of multiple upstream opacifying processes.

One edge warrants explicit comment: the relationship between pneumothorax and support devices is bidirectional, reflecting two distinct clinical realities. A pneumothorax may necessitate intercostal drain insertion, while central venous catheter placement can itself cause iatrogenic pneumothorax. We retain both directions in the graph and acknowledge that this constitutes a cycle; users of the SCM should select a single direction based on the causal query of interest.
\subsubsection{Continuous-Time Flow SCM Architecture}
\label{appsec:flow_scm_explanation}
%
There are 19 continuous-time flow MLP models in our metadata SCM, all trained concurrently under the joint flow matching objective in Eq \eqref{eq:fm_scm}. We propose a metadata conditioning strategy that remains consistent for all flows in the SCM, regardless of the number of causal parents each variable has.

Similar to the conditioning strategy described in Section \ref{subsec:flow_arch}, each variable in the causal graph is mapped to a learnable $d$-dimensional embedding. For categorical variables, this is done using a learned embedding table. For continuous variables (i.e. `Age'), we rescale and encode them with the same Fourier parameterisation used by \textcite{karras2024analyzing}, then project the result by an MLP to the shared embedding dimension $d$. Next, we define the parent index set for each node
\begin{align}
    &&\mathcal{I}^{(i)} \coloneqq \{j \in \{1, \ldots, n  \} \, : \, X^{(j)} \in \mathbf{PA}^{(i)} \}, && k^{(i)} = |\mathcal{I}^{(i)}| = |\mathbf{PA}^{(i)}|, && \text{for}~ i=1,\dots,n. &&
\end{align}
Given all $n=19$ learned embeddings, the conditioning signal $\mathbf{z}^{(i)} \in \mathbb{R}^d$ for each flow model in the SCM is given by a magnitude-preserving scaled sum of the parent embeddings of $X^{(i)}$:
\begin{align}
    && \mathbf{z}^{(i)} = \frac{1}{\sqrt{k^{(i)}}} \sum_{j \in \mathcal{I}^{(i)}} \mathbf{e}^{(j)}, && \mathbf{e}^{(j)} \coloneqq E\big(x^{(j)}; \phi^{(j)}\big) \in \mathbb{R}^d, \qquad j \in  \{1,\ldots,n\}. &&
\end{align}
The input to each flow MLP is then a concatenation of the time variable $t \sim \mathcal{U}(0,1)$, the flow interpolant value in one-hot space $y_t^{(i)} = (1-t)u^{(i)} + t\operatorname{onehot}(x^{(i)})$, and the respective parent conditioning embedding $\mathbf{z}^{(i)} \in \mathbb{R}^d$. For the MLPs, we stack 3 of the following residual blocks:
\begin{align}
    y^{(i)} \mapsto \textsc{Linear} \circ \textsc{SiLU} \circ \textsc{Dropout} \circ \textsc{Linear} \circ \textsc{LayerNorm}(\operatorname{onehot}{(x^{(i)})}),
\end{align}
with a 512 hidden dimension. Each MLP receives a non-linear projection of the concatenated inputs explained above, and each MLP's head has a \textsc{LayerNorm} and a final \textsc{Linear} layer to predict the velocity field of its respective flow. The final linear layer in each residual block is zero-initialised so that each block initially behaves as an identity transformation, which stabilises early training.
\paragraph{Samples, Interventions, and Counterfactuals.} At inference time, random samples from our continuous-time flow SCM are drawn by first sampling from the root node's source distribution and solving each subsequent ODE ancestrally in a \textit{topological} ordering of the associated causal graph. 

Let $(\pi_1,\ldots,\pi_n)$ denote a topological ordering of the variables $X_{1},\ldots,X_{n}$, such that for every edge $X_{i} \to X_{j}$ in the graph we have that $\pi^{-1}(i)<\pi^{-1}(j)$. In simple terms, $\pi^{-1}(i)<\pi^{-1}(j)$ means variable $X_i$ is processed before variable $X_j$, since $X_i$ causes $X_j$. We traverse the causal graph according to this topological ordering, and at step $k$ we process variable $i = \pi_k$.

For all categorical variables $Y_i = \operatorname{onehot}(X_i)\in \{0,1\}^K$ in turn, continuous ODE solver outputs 
\begin{align}
    && \hat{Y}_{i} = U_{i} + \int_0^1 v_i(t,Y_i(t); \mathbf{PA}_i) \, \mathrm{d}t, && U_{i} \sim p_{\text{src}}^{(i)}, && \hat{Y}_{i} \in \mathbb{R}^K,&&
\end{align}
are converted to discrete predictions via an argmax operation over $K$ categories\footnote{For simplicity, we slightly abuse notation by using $K$ to denote the number of categories for all categorical variables, although different variables may have different numbers of classes.}:
\begin{align}
    \label{eq:argmax}
    &&\hat{X}_{i} = \argmax_{k\in\{1,\ldots,K\}}\hat{Y}^k_{i}, && \text{for}~ i=1,\dots,n. &  &
\end{align}
Each of these discrete predictions is used to retrieve the corresponding entry from the embedding table, which then conditions the flows of its descendants, until the graph has been fully traversed.

Sampling from the \textit{interventional} distribution is similar, but with one or more observed variables intervened upon and detached from their parents. No ODE solving is needed for such variables.

Finally, to compute \textit{counterfactuals}, we use the \textit{abduction-action-prediction} \parencite{pearl2009causality} procedure for continuous-time flows following \textcite{ribeiro2025counterfactual}, but applied to our categorical variable case. In particular, \textcite{ribeiro2025counterfactual} used continuous flows for their image model only, relying on the classical Normalizing Flow approach of \textcite{pawlowski2020deep,ribeiro2023high} for their SCM, which does not support deterministic abduction for discrete variables. Our abduction step proceeds ancestrally in a topological ordering of the graph, inferring all posterior exogenous noise variables:
\begin{align}
    \hat{U}_{i} =Y_i - \int_0^1 v_i(t,Y_i(t); \mathbf{PA}_i) \, \mathrm{d}t \; \Rightarrow \; Y^{*}_i = \hat{U}_{i} + \int_0^1 v_i(t,Y_i{(t)}; \mathbf{PA}_i^{*}) \, \mathrm{d}t,
\end{align}
then the corresponding counterfactuals $Y^{*}_i$, under one or more interventions on the parents $\mathbf{PA}^{*}_i$.
Like before, continuous ODE solver outputs $Y^{*}_i(1) \in \mathbb{R}^K$ are converted to discrete counterfactual predictions $X^{*}_i$ via an argmax operation over $K$ categories (cf. Eq.~\eqref{eq:argmax}). 
As the graph is traversed in topological order, each predicted value is used to retrieve its corresponding embedding from the embedding table, which then conditions the flow vector fields of its causal descendants.
\section{Experiment Details \& Additional Results}
\label{appsec:experiments}
\subsection{Evaluation Metrics}
\label{app:metrics}

\paragraph{Distributional Fidelity Metrics.}
To evaluate the generative fidelity of our models, we measure the similarity between generated images and real test images using standard distributional fidelity and diversity metrics, but computed in multiple different feature spaces.
Fréchet Inception Distance (FID) \parencite{heusel2017gans} compares the feature distribution of real and generated images by approximating both distributions as multivariate Gaussians. 
Given feature means and covariances $(\mu_r, \Sigma_r)$ for real images and $(\mu_g, \Sigma_g)$ for generated images, FID is defined as:
\begin{align}
    \mathrm{FID}
    =
    \|\mu_r - \mu_g\|_2^2
    +
    \mathrm{Tr}(
        \Sigma_r + \Sigma_g
        - 2(\Sigma_r \Sigma_g)^{1/2}
    ).
\end{align}
Standard FID is computed using InceptionV3 features. 
In addition, we report analogous Fréchet distances in radiographic or self-supervised feature spaces, denoted FDD, by replacing InceptionV3 features with representations extracted from Rad-DINO or DINOv3. Kernel Inception Distance (KID) \parencite{binkowski2018demystifying} measures the squared maximum mean discrepancy between real and generated feature distributions. 
Similar to FDD, we denote the corresponding kernel distance computed in DINO-based feature spaces as KDD. 
Together, FID/FDD and KID/KDD provide complementary measures of whether generated samples match the real data distribution in both generic natural-image and domain-relevant radiographic representation spaces.

\textbf{Precision, Recall, Density, and Coverage.} \;
We further report Precision, Recall, Density, and Coverage \parencite{naeem2020reliable} to characterise different aspects of generative performance. 
Precision and Density measure the fidelity of generated samples by evaluating how well they lie on or near the real data manifold. 
Recall and Coverage measure sample diversity by evaluating how well the generated samples cover the support of the real data distribution. 

\textbf{Reconstruction Quality Metrics.} \;
For reconstruction quality, we report both pairwise image similarity metrics and distributional reconstruction fidelity. 
Structural Similarity Index Measure (SSIM) measures perceptual structural agreement between each input image and its reconstruction, while Peak Signal-to-Noise Ratio (PSNR) measures pixel-level reconstruction error. We also report Reconstruction Fréchet Distance (rFD), which applies the same Fréchet-distance formulation as FID, but compares the feature distribution of original images with that of their reconstructions rather than comparing real images with randomly generated samples. 
In our experiments, rFD is computed using Rad-DINO features rather than InceptionV3 features. 
Therefore, rFD evaluates whether the VAE reconstructions preserve the distribution of clinically relevant image features. 

\textbf{Patient Identity Preservation Metrics.} \;
For image editing, we further evaluate whether the model preserves image content that should remain unchanged under the intervention. 
We refer to this property as patient identity preservation. 
Given an original image $\mathbf{x}$ and its edited counterpart $\mathbf{x}^*$, we measure identity preservation using complementary pixel-space, global feature-space, and attribute-specialised feature-space metrics. First, we compute SSIM between $\mathbf{x}$ and $\mathbf{x}^*$ in pixel space. 
While SSIM is also used for reconstruction evaluation, here it measures whether the target edit preserves low-level anatomical and acquisition-specific image structure. 
Higher SSIM indicates stronger structural preservation. Second, we compute LPIPS-style perceptual distances in pretrained representation spaces. 
Unlike standard LPIPS, which is typically computed using natural-image backbones, we compute feature-space distances using Rad-DINO and DINOv3 representations. 
Given a feature extractor $\phi$, let $\phi_\ell(\mathbf{x}) \in \mathbb{R}^{N_\ell \times C_\ell}$ denote the hidden representation of image $\mathbf{x}$ at layer $\ell$, where $N_\ell$ is the number of tokens and $C_\ell$ is the feature dimension. 
For a set of selected layers $\mathcal{L}$, we define the perceptual distance between the original image $\mathbf{x}$ and its edit $\mathbf{x}^*$ as
\begin{align}
    d_{\phi}(\mathbf{x}, \mathbf{x}^*)
    =
    \sum_{\ell \in \mathcal{L}}
    \frac{1}{N_\ell C_\ell}
    \left\|
        \tilde{\phi}_\ell(\mathbf{x})
        -
        \tilde{\phi}_\ell(\mathbf{x}^*)
    \right\|_F^2 ,
\end{align}
where $\tilde{\phi}_\ell(\cdot)$ denotes the layer representation after optional feature normalization, and $\|\cdot\|_F$ is the Frobenius norm over token and feature dimensions. 
Lower values indicate that the edited image remains closer to the original image in semantic feature space, and therefore better preserves identity-relevant visual content. We also report a relative version of this perceptual distance, which normalises the edit-induced feature distance by the average distance between randomly paired real images:
\begin{align}
    d_{\phi}^{\mathrm{rel}}(\mathbf{x}, \mathbf{x}^*)
    =
    \frac{
        d_{\phi}(\mathbf{x}, \mathbf{x}^*)
    }{
        \frac{1}{M}
        \sum_{i=1}^{M}
        d_{\phi}(\mathbf{x}'_i, \mathbf{x}''_i)
    } ,
\end{align}
where $\{(\mathbf{x}'_i, \mathbf{x}''_i)\}_{i=1}^{M}$ denotes a set of randomly sampled image pairs. 
This relative score expresses the magnitude of the edit change as a fraction of the typical feature-space distance between unrelated images. 
It therefore makes identity preservation scores easier to interpret and compare across feature extractors such as Rad-DINO and DINOv3. Finally, to assess whether non-intervened patient attributes are preserved, we compute cosine similarity in attribute-specialised latent spaces. 
Let $\psi_a(\cdot)$ denote the embedding extracted from the prediction head corresponding to attribute $a$, and let $\mathcal{A}$ denote the set of attributes left unchanged by the edit. We then compute
\begin{align}
    s_a(\mathbf{x}, \mathbf{x}^*)
    =
    \frac{1}{|\mathcal{A}|}
    \sum_{a \in \mathcal{A}}
    \frac{
        \psi_a(\mathbf{x})^\top \psi_a(\mathbf{x}^*)
    }{
        \|\psi_a(\mathbf{x})\|_2
        \|\psi_a(\mathbf{x}^*)\|_2
    } .
\end{align}
Higher attribute-space cosine similarity indicates stronger preservation of metadata-relevant information that should remain unchanged after editing. Together, these metrics provide complementary views of identity preservation.
SSIM measures low-level structural similarity, Rad-DINO and DINOv3 LPIPS-style distances measure global semantic preservation, and attribute-specialised cosine similarity measures whether invariant patient attributes remain stable under editing.
\subsection{Improved VAEs for Radiographic Perceptual Quality}
\label{appsec:experiments_radvae}
\begin{table}[!ht]
    \small
    \centering
    \caption{Training hyperparameters for the proposed RadVAE trained from scratch on CXR7-1M.}
    \vspace{2pt}
    \begin{tabular}{lr|lr}
        \toprule
        \multicolumn{4}{c}{\textbf{RadVAE Training}} \\[2pt]
        Setting & Value & Setting & Value \\
        \midrule
        Optimizer & AdamW & Batch size & 192 \\
        Learning rate & 1e-4 & LR warmup steps & 4000 \\
        Weight decay & 1e-4 & Adam betas & [0.9, 0.999] \\
        EMA rate & 0.9999 & KL weight ($\beta$) & 0.0 \\
         DINO version & Rad & Rad-LPIPS weight ($\alpha$) & 250 \\
         Num. nodes & 4 & Num. GPUs & 16 \\
        \bottomrule
    \end{tabular}
    \label{apptab:rad_vae_train}
\end{table}
Table \ref{apptab:rad_vae_train} reports the training hyperparameters we used to train our RadVAE. The model was trained for 630K steps using multi-node distributed data parallelism in bfloat16 precision, on 16 NVIDIA GH200 GPUs split over 4 compute nodes \parencite{mcintoshsmith2024isambardai}. In all cases, model development and training were performed using PyTorch \parencite{paszke2019pytorch}.

\begin{table}[!t]
    \footnotesize
    \centering
    \caption{
    \textbf{Ablation study of VAE radiographic fine-tuning strategies}. We fine-tune (FT) the FLUX.2 VAE with different Rad-DINO perceptual loss and LoRA setups, closing the radiographic fidelity gap to our baseline RadVAE trained from scratch, whilst outperforming it on all other metrics. The blue row reports the baseline FLUX.2 VAE performance on the CXR7-1M test set when simply averaging decoder RGB predictions; remaining rows show results from RadVAE{\scriptsize\textcolor{steelblue}{$^\text{FLUX.2}$}} fine-tuning variants.
    }
    \label{apptab:vae_results}
    \renewcommand{\arraystretch}{1.05}
    \vspace{2pt}
    \begin{tabular}{lccccc}
        \toprule
         \multirow{2}{*}{\textbf{Model}} & \multicolumn{2}{c}{\textbf{Reconstruction FD} $\downarrow$ } & \multirow{2}{*}{\textbf{PSNR} $\uparrow$} & \multirow{2}{*}{\textbf{SSIM} $\uparrow$} \\
         & \footnotesize{Rad-DINO} & \footnotesize{DINOv3} & & \\
        \midrule
        RadVAE (Scratch) & \textbf{0.0476} & 2.246 & 43.706 & 0.9831 \\
        \midrule
        \rowcolor{babyblueice!20}
        \hspace{0pt}FLUX.2 (RGB-mean) & 0.0887 & 0.6320 & 45.413 & 0.9875 \\
        \hspace{4pt}Luma Coefficients & 0.0903 & 0.6231 & 45.699 & 0.9878 \\
        \hspace{4pt}LoRA (Mid-Block) & 0.0755 & 0.5962 & 45.352 & 0.9881 \\
        \hspace{4pt}FT RGB Head & 0.0994 & 0.6019 & 45.552 & 0.9875 \\
        \hspace{4pt}+ Rad-DINO Perceptual & 0.0616 & 0.6657 & 45.555 & 0.9872 \\
        \hspace{4pt}\, \, + FT Last Block & 0.0722 & 0.8946 & \textbf{45.905} & 0.9869 \\
        \hspace{4pt}+ LoRA (Mid-Block, $r{=}$16, $\alpha{=}$16) & 0.0574 & 0.6527  &  45.260 & 0.9862 \\
        \hspace{4pt}+ LoRA (Mid-Block, $r{=}$8, $\alpha{=}$8) & 0.0539 & 0.6257  &  45.413 & 0.9861 \\
        \hspace{4pt}+ LoRA (Up-Blocks, $r{=}$8, $\alpha{=}$8) & \underline{0.0487} & 0.5933 & 45.829 & 0.9880 \\
        \hspace{4pt}+ LoRA (Up-Blocks, $r{=}$8, $\alpha{=}$16) & 0.0532 & \textbf{0.5775} & 45.867 & \textbf{0.9882} \\
        \bottomrule
    \end{tabular}
\end{table}
\textbf{Radiographic Perceptual Training.} \; As mentioned in \S\ref{subsec:vae_exp}, we find that averaging decoder RGB predictions achieves highly competitive results on CXR when using the FLUX.2 VAE in particular. Since the exact training recipe of the FLUX.2 VAE has not been publicly released at the time of this writing, it is difficult to attribute this performance to any specific design or training factor. That said, the FLUX.2 VAE still underperforms our baseline RadVAE trained from scratch in terms of radiographic fidelity (Rad-DINO rFD). To close this gap, we explored several different radiographic perceptual fine-tuning setups using Rad-DINO \parencite{perez2025exploring} and LoRA \parencite{hu2022lora}. We elected to tune only the FLUX.2 VAE decoder to avoid affecting the pretrained latent space, particularly since the original training recipe is not public. Training only the decoder improves radiographic reconstruction quality while preserving compatibility with the pretrained encoder. All our fine-tuned models were trained on CXR7-1M and ran for a maximum 50K steps, as early experiments showed that performance saturated relatively quickly. We fine-tuned using the Adam optimiser with a learning rate of 2e-5, batch size of 16, 500 linear warmup steps, no weight decay, Adam betas of [0.9, 0.999], and an EMA rate of 0.995. We ran a light sweep over the Rad-DINO perceptual loss weight, using $\alpha \in \{10, 100, 200, 500, 1000\}$, and found $\alpha = 500$ to perform best. We selected the checkpoint with the lowest validation loss for evaluation on the CXR7-1M test set.

The results of our ablation study are reported in Table \ref{apptab:vae_results}. The first thing we tried that didn't work well enough was replacing the simple RGB-mean FLUX.2 decoder prediction with a weighted prediction by Luma coefficients: (0.29894, 0.58704, 0.11402). Luma coefficients weight the RGB channels according to perceived brightness, giving more weight to green and less to blue. This slightly improved all reconstruction metrics except for the intended Rad-DINO rFD. We then evaluated several decoder fine-tuning strategies with and without a Rad-DINO perceptual loss: full decoder fine-tuning, RGB head only fine-tuning, joint fine-tuning of the RGB head and the final decoder block, and LoRA fine-tuning of selected mid-block and up-block layers. We found that full decoder fine-tuning led to undesirable overfitting. 
The best approach was a radiographic perceptual LoRA configuration that updated the decoder input convolution, selected mid-block residual and attention projections, one residual convolution from each upsampling block, and fine-tuned the RGB head.

\subsection{Evaluating Generative Fidelity: Benchmarks \& Clinical Expert Study}
\label{appsec:generative_fidelity}

Tables \ref{apptab:radit_training} and \ref{apptab:radunet_training} report the training hyperparameters we used to train our transformer and U-Net-based flow models, respectively. Training was performed in PyTorch \parencite{paszke2019pytorch}. For the transformer models, hyperparameters were largely adopted from prior work, with minor adjustments based on initial experiments. In particular, we found that learning-rate warmup degraded training stability early on, so we disabled it. To facilitate comparison, the base transformer models (RadiT B) were trained with identical training hyperparameters, except that the pixel-space version used a shifted timestep schedule $t \mapsto t /(\alpha - t(\alpha -1))$ with $\alpha=3$, as prescribed by \parencite{esser2024scaling}. Intuitively, larger $\alpha$'s compensate for the increased redundancy at higher resolutions and can boost performance.
Since our pixel-space RadUNet model departs from the original EDM2 configuration presets \parencite{karras2024analyzing}, partly due to compute constraints, we took a more conservative training setup (e.g. reduced the learning rate to 1e-4). As for the latent-space RadUNet, we used the EDM2-M setup and adjusted the hyperparameters for a 1M-step run to match our compute resources. Due to using a smaller batch size, we reduced the learning rate to 1e-2, with 50K warmup steps. 

We elect RadiT XL (latent) and RadUNet (pixel) as the most capable variants from their respective model classes. For all our generative fidelity results, we used a \texttt{dopri5} ODE solver \parencite{torchdiffeq} with \texttt{atol} and \texttt{rtol} of 1e-5, with identical seed conditions. Tables \ref{tab:sample_dinov3} and \ref{tab:sample_inception} report additional results using DINOv3 \parencite{simeoni2025dinov3} and Inceptionv3 \parencite{szegedy2016rethinking} as feature extractors, respectively. We also conducted a reference split sensitivity test, and the results are reported in Table \ref{apptab:split_ablation}. Since part of CheXGenBench overlaps with the CXR7-1M training split, we repeated the FDD evaluation using three 5K reference splits sampled from the CXR7-1M train and test splits. The results are similar between CXR7-1M train and test references, showing that the measured fidelity is not primarily driven by train-set overlap. The larger shift relative to ChexGenBench, especially for Rad-DINO FDD, indicates that absolute FDD values depend on the chosen reference distribution.
\begin{table}[!t]
    \small
    \centering
    \caption{
    \textbf{Comparative evaluation of CXR generative fidelity.} 
    All metrics were computed using DINOv3 features.   
    Benchmark results are from \textit{CheXGenBench} \parencite{dutt2025chexgenbench}. We also report results on two internal test splits from \textit{CXR7-1M}, a MIMIC-CXR 5K split and a separate 50K split.
    Superscript ($^{\textcolor{steelblue}{\text{pix}}}$) denotes our flow model variants trained in pixel-space ($512{\times}512$ resolution). 
    }
    \label{tab:sample_dinov3}
    \vspace{2pt}
    \begin{tabular}{lccccccr}
        \toprule
         \multirow{2}{*}{\textbf{Model}} & \textbf{FDD} $\downarrow$ & \textbf{KDD} $\downarrow$& \textbf{Precision} $\uparrow$ & \textbf{Recall} $\uparrow$ & \textbf{Density} $\uparrow$ & \textbf{Coverage} $\uparrow$ & \multirow{2}{*}{\textbf{Size}} \\
         & \tiny{(DINOv3)} & \tiny{(DINOv3)} & \tiny{(DINOv3)} & \tiny{(DINOv3)} & \tiny{(DINOv3)} & \tiny{(DINOv3)} &  \\
        \midrule
\rowcolor{babyblueice!20}\hspace{-4pt}\textit{CheXGenBench} & & & & & & & \\
 RadiT B$^{\textcolor{steelblue}{\text{pix}}}$ & 9.6176 & 0.0398 & 0.3776 & 0.1170
 & 0.2004
 & 0.2733
 & 0.3B \\
 RadiT B & 1.7651 & 0.0050 & 0.7795 & 0.6662 
 & 0.5702
 & 0.7243
 & 0.3B \\
 RadUNet$^{\textcolor{steelblue}{\text{pix}}}$ & 5.6389 & 0.0210 & 0.5638 & 0.2243
 & 0.4546
 & 0.4871
 & 0.3B \\
 RadUNet & 2.8031 & 0.0095 & 0.7906 & 0.6486 
 & 0.4734
 & 0.5186
 & 0.5B \\
\textbf{RadiT XL} & 1.2907 & 0.0013 & 0.7590 
  & 0.8602 
 & 0.5683
 & 0.6369
 & 1.3B \\
 \midrule
\rowcolor{babyblueice!20}\hspace{-4pt}\textit{CXR7-1M (MIMIC)} & & & & & & & \\
 RadiT B$^{\textcolor{steelblue}{\text{pix}}}$ & 10.066 & 0.0385 & 0.4811 & 0.1444
 & 0.2854
 & 0.3828
 & 0.3B \\
     RadiT B & 1.6791 & 0.0042 & 0.7614 & 0.6428
 & 0.8288
 & 0.8933
 & 0.3B \\
 RadUNet$^{\textcolor{steelblue}{\text{pix}}}$ & 5.6100 & 0.0199 & 0.6820 & 0.2644
 & 0.5991
 & 0.6351
 & 0.3B \\
  RadUNet & 2.5850 & 0.0079 & 0.7396 & 0.5159
 & 0.7537
 & 0.8343
 & 0.5B \\
  \textbf{RadiT XL} & 1.3142 & 0.0033 & 0.7749
  & 0.7040
 & 0.8471
 & 0.9178
 & 1.3B \\
 \midrule
\rowcolor{babyblueice!20}\hspace{-4pt}\textit{CXR7-1M (50K)} & & & & & & & \\
    \textbf{RadiT XL} & 0.8342 & 0.0031 & 0.7477
  & 0.6772
 & 0.7960
 & 0.8806
 & 1.3B \\
         \bottomrule
    \end{tabular}
\end{table}
\begin{table}[!ht]
    \small
    \centering
    \caption{
    \textbf{Comparative evaluation of CXR generative fidelity.} 
    All metrics were computed using Inceptionv3 features.   
    Benchmark results are from \textit{CheXGenBench} \parencite{dutt2025chexgenbench}. We also report results on two internal test splits from \textit{CXR7-1M}, a MIMIC-CXR 5K split and a separate 50K split.
    Superscript ($^{\textcolor{steelblue}{\text{pix}}}$) denotes our flow model variants trained in pixel-space ($512{\times}512$ resolution). 
    }
    \label{tab:sample_inception}
    \vspace{2pt}
    \begin{tabular}{lccccccr}
        \toprule
         \multirow{2}{*}{\textbf{Model}} & \textbf{FDD} $\downarrow$ & \textbf{KDD} $\downarrow$& \textbf{Precision} $\uparrow$ & \textbf{Recall} $\uparrow$ & \textbf{Density} $\uparrow$ & \textbf{Coverage} $\uparrow$ & \multirow{2}{*}{\textbf{Size}} \\
         & \tiny{(Inceptionv3)} & \tiny{(Inceptionv3)} & \tiny{(Inceptionv3)} & \tiny{(Inceptionv3)} & \tiny{(Inceptionv3)} & \tiny{(Inceptionv3)} &  \\
        \midrule
\rowcolor{babyblueice!20}\hspace{-4pt}\textit{CheXGenBench} & & & & & & & \\
 RadiT B$^{\textcolor{steelblue}{\text{pix}}}$ & 10.870 & 0.0082 & 0.7688 & 0.6381
 & 0.8430
 & 0.8198
 & 0.3B \\
 RadiT B & 4.7431 & 0.0011 & 0.8053 & 0.7437
 & 1.0350
 & 0.9499
 & 0.3B \\
  RadUNet$^{\textcolor{steelblue}{\text{pix}}}$ & 7.8100 & 0.0045 & 0.7912 & 0.6855
 & 0.9834
 & 0.8953
 & 0.3B \\
 RadUNet & 5.3172 & 0.0017 & 0.8166 & 0.7193
 & 1.0773
 & 0.9478
 & 0.5B \\
 \textbf{RadiT XL} & 4.2102 & 0.0006 & 0.8063 
  & 0.7775
 & 0.9839
 & 0.9507
 & 1.3B \\
 \midrule
\rowcolor{babyblueice!20}\hspace{-4pt}\textit{CXR7-1M (MIMIC)} & & & & & & & \\
 RadiT B$^{\textcolor{steelblue}{\text{pix}}}$ & 9.4911 & 0.0056 & 0.7791 & 0.6615
 & 0.8924
 & 0.8826
 & 0.3B \\
     RadiT B & 4.6394 & 0.0006 &  0.8133 & 0.7473
 & 1.0871
 & 0.9682
 & 0.3B \\
 RadUNet$^{\textcolor{steelblue}{\text{pix}}}$ & 6.9954 & 0.0031 & 0.8127 & 0.6980
 & 1.0734
 & 0.9331
 & 0.3B \\
  RadUNet & 5.2601 & 0.0011 & 0.8208 & 0.7277
 & 1.0992
 & 0.9585
 & 0.5B \\
 \textbf{RadiT XL} & 4.2529 & 0.0004 & 0.8186
  & 0.7692
 & 1.0631
 & 0.9664
 & 1.3B \\
 \midrule
\rowcolor{babyblueice!20}\hspace{-4pt}\textit{CXR7-1M (50K)} & & & & & & & \\
    \textbf{RadiT XL} & 0.8668 & 0.0004 & 0.8062
  & 0.7507
 & 1.0566
 & 0.9635
 & 1.3B \\
         \bottomrule
    \end{tabular}
\end{table}

\begin{table}[!ht]
    \small
    \centering
    \caption{Transformer training hyperparameters for our CXR flow models (RadiT). 
    }
    \vspace{2pt}
    \begin{tabular}{l|ccc}
        \toprule
        \multirow{2}{*}{Setting} & \textbf{RadiT B}$^{\textcolor{steelblue}{\text{pix}}}$ & \textbf{RadiT B} & \textbf{RadiT XL} \\
        & (Pixel)  & (Latent) & (Latent) \\
        \midrule
        Optimizer & Adam & Adam & Adam \\
        Learning rate & 1e-4 & 1e-4 & 1e-4 \\
        LR warmup steps & 0 & 0 & 0 \\
        Batch size & 72 & 72 & 512 \\
        Weight decay & 0 & 0 & 0 \\
        Adam betas & [0.9, 0.999] & [0.9, 0.999] & [0.9, 0.999]  \\
        Gradient clipping & 1.0 & 1.0 & 1.0 \\
        Shifted schedule $\alpha$ & 3.0 & 1.0 & 1.0 \\
        Uniform dequant. & Yes & No & No \\
        EMA rate & 0.9999 & 0.9999 & 0.9999 \\
        Training steps & 1M & 1M & 3M  \\
        Training tokens & 73.7B & 73.7B & 1.57T \\
        Tokens/s & 100K & 200K & 1M \\
        Num. nodes & 1 & 1 & 8  \\
        Num. GPUs & 2 & 2 & 32  \\
        Precision & bfloat16 & bfloat16 & bfloat16 \\
        \bottomrule
    \end{tabular}
    \label{apptab:radit_training}
\end{table}

\begin{table}[!ht]
    \small
    \centering
    \caption{UNet training hyperparameters for our CXR flow models (RadUNet). 
    }
    \vspace{2pt}
    \begin{tabular}{l|cc}
        \toprule
        \multirow{2}{*}{Setting} & \textbf{RadUNet}$^{\textcolor{steelblue}{\text{pix}}}$ & \textbf{RadUNet} \\
        & (Pixel)  & (Latent) \\
        \midrule
        Optimizer & AdamW & Adam  \\
        Learning rate & 1e-4 & 1e-2 \\
        LR warmup steps & 4000 & 50K \\
        Batch size & 72 & 72 \\
        Weight decay & 1e-4 & 0 \\
        Adam betas & [0.9, 0.99] & [0.9, 0.99] \\
        Gradient clipping & 1.0 & 1.0 \\
        Shifted schedule $\alpha$ & 3 & 1 \\
        Uniform dequant. & Yes & No \\
        EMA rate & 0.9999 & 0.9999 \\
        Training steps & 2M & 1M \\
        Num. nodes & 1 & 1   \\
        Num. GPUs & 6 & 2  \\
        Precision & bfloat16 & bfloat16 \\
        \bottomrule
    \end{tabular}
    \label{apptab:radunet_training}
\end{table}

\begin{table}[!ht]
    \small
    \centering
    \caption{
    \textbf{Reference split sensitivity of FDD estimates.} We compare the CheXGenBench \parencite{dutt2025chexgenbench} reference split against internal 5K-sample reference subsets drawn from CXR7-1M train and test splits. Reporting RadiT XL FDD across 3 seeds using Rad-DINO, DINOv3, and Inceptionv3.}
    \label{apptab:split_ablation}
    \vspace{2pt}
    \begin{tabular}{llccc}
         \toprule
         \multirow{2}{*}{\textbf{Model}} & \multirow{2}{*}{\textbf{Ref. Split} (5K)} & \textbf{FDD} & \textbf{FDD} & \textbf{FDD} \\
           &  & \footnotesize{(Rad-DINO)} & \footnotesize{(DINOv3)} & \footnotesize{(Inceptionv3)} \\
           \midrule
         RadiT XL & CheXGenBench & 13.152 & 1.2907 & 4.2102 \\
         \midrule
         RadiT XL & CXR7-1M (Train) & 8.367$\pm$0.081 & 1.356$\pm$0.035 & 4.228$\pm$0.015\\
         RadiT XL & CXR7-1M (Test) & 9.188$\pm$0.045 &  1.356$\pm$0.040 & 4.241$\pm$0.011 \\
        \bottomrule
    \end{tabular}
\end{table}

\begin{table}[!t]
    \small
    \centering
    \caption{
    \textbf{Quantitative evaluation of identity preservation under controlled editing.} We report SSIM and LPIPS loss between observed CXR images and their edited counterparts, with the latter computed using both Rad-DINO and DINOv3. Since LPIPS losses are naturally small in magnitude, we also report \textit{relative} LPIPS (rLPIPS) with respect to a baseline loss on 5K random image pairings. Although RadiT XL is by far the superior generative model in terms of fidelity (cf. Table \ref{tab:sample_results}), we find that it is outperformed by the pixel-space RadUNet$^{\textcolor{steelblue}{\text{pix}}}$ in terms of identity preservation under editing. Overall, we find that pixel-space models consistently rank top in terms of identity preservation. It's important to note that improved identity preservation alone does not imply a better model, as a perfect score can be attained by not applying any changes. Thus, practitioners should consider the trade-off between edit effectiveness and identity preservation when selecting models for downstream tasks.
    }
    \label{apptab:identity_pres}
    \vspace{2pt}
    \begin{tabular}{llccccc}
         \toprule
         \multirow{2}{*}{\textbf{Model}} & \multirow{2}{*}{\textbf{Edit}} & \multirow{2}{*}{\textbf{SSIM} $\uparrow$} & \textbf{rLPIPS} $\downarrow$ & \textbf{LPIPS} $\downarrow$ & \textbf{rLPIPS}$\downarrow$ & \textbf{LPIPS} $\downarrow$ \\
          &  &  & \footnotesize{(Rad-DINO)} & \footnotesize{(Rad-DINO)} & \footnotesize{(DINOv3)} & \footnotesize{(DINOv3)} \\
         \midrule
         Random Pairs & N/A & 0.4780 & 1.0 & 0.01020 & 1.0 & 0.01155 \\
         \midrule
         {RadUNet}$^{\textcolor{steelblue}{\text{pix}}}$ & Age & {0.8813} & {0.2559} & {0.00261} & {0.3030} & {0.00350} \\
         RadiT B$^{\textcolor{steelblue}{\text{pix}}}$ & Age & 0.8657 & 0.2851 & 0.00291 & 0.3299 & 0.00381 \\
         RadiT B & Age & 0.8527 & 0.2866 & 0.00292 & 0.3295 & 0.00381 \\
         RadiT XL & Age & 0.8255 & 0.3471 & 0.00354 & 0.3948 &  0.00456 \\
         RadUNet & Age & 0.8109 & 0.3726 & 0.00380 & 0.4097 & 0.00473 \\
         \midrule
         {RadUNet}$^{\textcolor{steelblue}{\text{pix}}}$ & Dataset & {0.6959} & {0.6048} & {0.00617} & {0.6295} & {0.00727} \\
        RadiT B$^{\textcolor{steelblue}{\text{pix}}}$ & Dataset & 0.6498 & 0.6939 & 0.00708 & 0.7115 & 0.00822 \\
         RadiT B & Dataset & 0.6051 & 0.7567 & 0.00772 & 0.7634 &  0.00882 \\
         RadiT XL & Dataset & 0.5975 & 0.7753 & 0.00791 & 0.7841 &  0.00906 \\
         RadUNet & Dataset & 0.5845 & 0.8004 & 0.00816 & 0.7997 &  0.00924 \\
         \midrule
         {RadUNet}$^{\textcolor{steelblue}{\text{pix}}}$ & Race &  {0.9283} & {0.1548} & {0.00158} & {0.2002} & {0.00231} \\
        RadiT B$^{\textcolor{steelblue}{\text{pix}}}$ & Race &  0.9203 & 0.1705 & 0.00174 & 0.2117 & 0.00245 \\
         RadiT B & Race & 0.9187 & 0.1517 & 0.00155 & 0.1940 & 0.00224 \\
         RadiT XL & Race & 0.8887 & 0.2179 & 0.00222 & 0.2713 & 0.00313 \\
         RadUNet & Race & 0.8712 & 0.2502 & 0.00255 & 0.2909 & 0.00336 \\
        \midrule
         {RadUNet}$^{\textcolor{steelblue}{\text{pix}}}$ & Sex &  {0.8734} & {0.2976} & {0.00304} & {0.3628} & {0.00419} \\
        RadiT B$^{\textcolor{steelblue}{\text{pix}}}$ & Sex &  0.8492 & 0.3410 & 0.00348 & 0.4004 & 0.00462 \\        
         RadiT B & Sex & 0.8272 & 0.3577 & 0.00365 & 0.4070 & 0.00470 \\
         RadiT XL & Sex & 0.8054 & 0.4088 & 0.00417 & 0.4602 &  0.00532 \\
         RadUNet & Sex & 0.7794 & 0.4583 & 0.00468 & 0.4984 & 0.00576 \\
         \midrule
         {RadUNet}$^{\textcolor{steelblue}{\text{pix}}}$ & View &  {0.6957} & {0.6305} & {0.00643} & {0.6623} & {0.00765} \\
        RadiT B$^{\textcolor{steelblue}{\text{pix}}}$ & View &  0.6459 & 0.7214 & 0.00736 & 0.7481 & 0.00864 \\
         RadiT XL & View & 0.6047 & 0.7833 & 0.00799 & 0.8023 &  0.00927 \\
         RadiT B & View & 0.6006 & 0.7860 &  0.00802 & 0.8009 & 0.00925 \\
         RadUNet & View & 0.5784 & 0.8296 &  0.00846 & 0.8367 & 0.00966 \\
         \midrule
        {RadUNet}$^{\textcolor{steelblue}{\text{pix}}}$ & Finding &  {0.9523} & {0.1067} & {0.00109} & {0.1372} & {0.00158} \\
         RadiT B & Finding & 0.9399 & 0.1108 & 0.00113 & 0.1454 & 0.00168 \\
         RadiT B$^{\textcolor{steelblue}{\text{pix}}}$ & Finding & 0.9264 & 0.1559 & 0.00159 &  0.1876 & 0.00217 \\
         RadiT XL & Finding & 0.9060 & 0.1825 & 0.00186 & 0.2264 & 0.00262 \\
         RadUNet & Finding & 0.8911 & 0.2092 & 0.00213 & 0.2417 & 0.00279 \\
         \bottomrule
    \end{tabular}
\end{table}
\subsection{Controllable Image Generation: Subgroup Analysis}
\label{appsec:experiments_control}
\begin{figure*}[!ht]
    \centering
    \includegraphics[trim=0 0 17 15,clip,width=\textwidth]{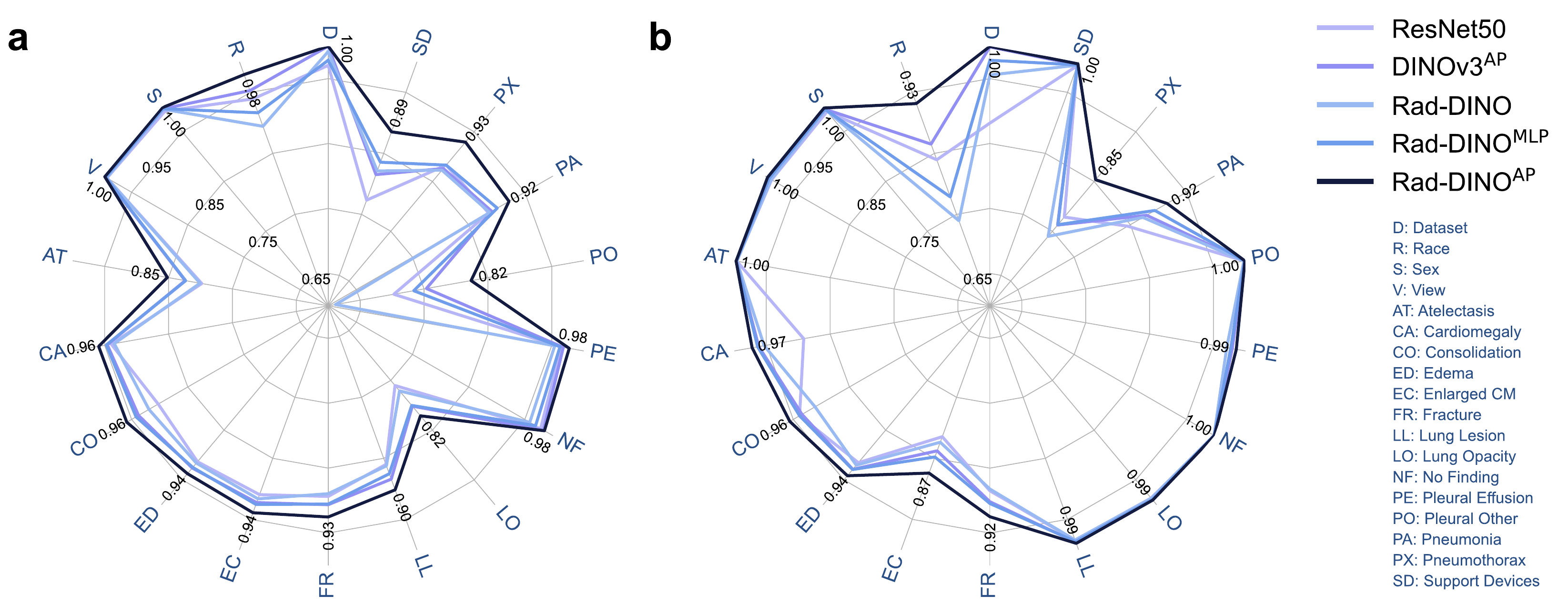}
    \caption{
    \textbf{Patient metadata and clinical finding predictors trained on CXR7-1M.} Radar plots comparing ResNet50, attention-pooled DINOv3$^{\text{AP}}$, and three Rad-DINO variants. Reporting ROCAUC on the left, and AUPRC on the right. We find that Rad-DINO$^{\text{AP}}$ outperforms all baselines, and given its strong performance, we expect it to be broadly useful beyond image editing evaluation.
    }
    \label{appfig:rad_dino_ap}
\end{figure*}
\textbf{Patient Metadata \& Clinical Finding Predictors.} \; The hyperparameters used for training our patient metadata predictors are reported in Table \ref{apptab:predictor_training}. As mentioned in the main text, all our models were trained to predict the 19 metadata variables available in CXR7-1M. We used a multi-task training approach, with individual Cross-Entropy losses for each categorical variable and an MSE loss for `Age'. We handle missing/ambiguous labels (`NaN') by simply skipping them in the loss computation. No extensive hyperparameter tuning was necessary to obtain strong performance. We experimented with standard data augmentation strategies (e.g. random flips, cropping, affine transformations, elastic deformations, Gaussian blur etc) but found that they did not meaningfully improve performance. Therefore, our final models were trained without any data augmentation for simplicity.
\begin{table}[!ht]
    \small
    \centering
    \caption{Training hyperparameters for patient metadata and clinical finding predictors. 
    }
    \vspace{2pt}
    \begin{tabular}{l|ccccc}
        \toprule
        Setting & \textbf{ResNet-50} & \textbf{DINOv3$^{\text{AP}}$} & \textbf{Rad-DINO} & \textbf{Rad-DINO$^{\text{MLP}}$} & \textbf{Rad-DINO$^{\text{AP}}$} \\
        \midrule
        Optimizer & AdamW & AdamW & AdamW & AdamW & AdamW  \\
        Learning rate & 1e-4 & 1e-4 & 1e-4 & 1e-4 & 1e-4  \\
        LR warmup steps & 100 & 100  & 100  & 100 & 100   \\
        Batch size & 128 & 360 & 512 & 512 & 360  \\
        Weight decay & 1e-4 & 1e-4 & 1e-4 & 1e-4 & 1e-4 \\
        Adam betas & [0.9, 0.999] & [0.9, 0.999] & [0.9, 0.999] & [0.9, 0.999] & [0.9, 0.999] \\
        Gradient clipping & 1.0 & 1.0 & 1.0 & 1.0 & 1.0  \\
        EMA rate & 0.9999 & 0.999 & 0.9999 & 0.9999 & 0.999 \\
        Hidden dim. size & n/a & n/a & n/a & 1024 & n/a \\
        Dropout & n/a & n/a & n/a & 0.1 & n/a \\
        Attn. pooling layers & n/a & [3, 6, 9, 12] & n/a & n/a & [3, 6, 9, 12] \\
        Max. training steps & 100K & 25K & 75K & 75K & 25K \\
        Num. nodes & 1 & 1 & 1 & 1 & 1    \\
        Num. GPUs & 1 & 2 & 2 & 2 & 2   \\
        Precision & bfloat16 & bfloat16 & bfloat16 & bfloat16 & bfloat16  \\
        \bottomrule
    \end{tabular}
    \label{apptab:predictor_training}
\end{table}

\textbf{Image Editing.} \; A randomly selected subset of 5K patients from the test split of CXR7-1M was used as the baseline for evaluating the editing effectiveness of our models. For our edited samples, we used the unique metadata profiles (the 19 target variables, including patient demographics, image acquisition characteristics, and clinical findings) from these same 5K patients to generate images. To perform an edit on an image, we first change the existing attribute value (\textit{e.g.}, Dataset $=$ BRAX, or Age $=$ 35) to a random value that is different from the initial one (\textit{e.g.}, Dataset $=$ VinDR-CXR, or Age $=$ 57). Based on this edit, we solve our Flow SCM model (with \texttt{dopri5}, \texttt{atol}${=}$1e-6, \texttt{rtol}${=}$1e-6) to predict the rest of the metadata profile such that it adheres to the causal graph in Figure \ref{fig:causal_graph}. Then, with the fully updated metadata profile based on the targeted edit, we perform the counterfactual inference steps of abduction, action, and prediction using our generative models as described in \S\ref{appsec:flow_scm_explanation}, with \texttt{dopri5} solver settings of \texttt{atol}$=$1e-5 and \texttt{rtol}$=$1e-5. 
We use a batch size of 32 for RadiT XL, RadUNet, and RadUNet$^{\textcolor{steelblue}{\text{pix}}}$, and a batch size of 16 for RadiT B and RadiT B$^{\textcolor{steelblue}{\text{pix}}}$.

\textbf{Identity Preservation.} \;
The main quantitative metrics we use for measuring patient identity preservation under image editing are SSIM and LPIPS loss. These are computed by comparing real observed CXR images with their edited counterparts. Rather than using VGG to compute the LPIPS loss \parencite{zhang2018unreasonable}, we use modern and more capable feature extractors, namely Rad-DINO and DINOv3 \parencite{dino_perceptual}. SSIM captures structural similarity, while LPIPS uses deep perceptual features and is generally more aligned with human judgements \parencite{zhang2018unreasonable}.

For our attribute-specialised embedding analysis, we utilise the task-conditioned attention-pooled embeddings from Rad-DINO$^{\text{AP}}$ for sex, race, age, view, and dataset prediction. We extract these 3072-dimensional representations for the original images, along with the edited images, and compute cosine similarity for all 5K pairs of images.   

\begin{figure}
    \centering
    \includegraphics[width=.999\textwidth]{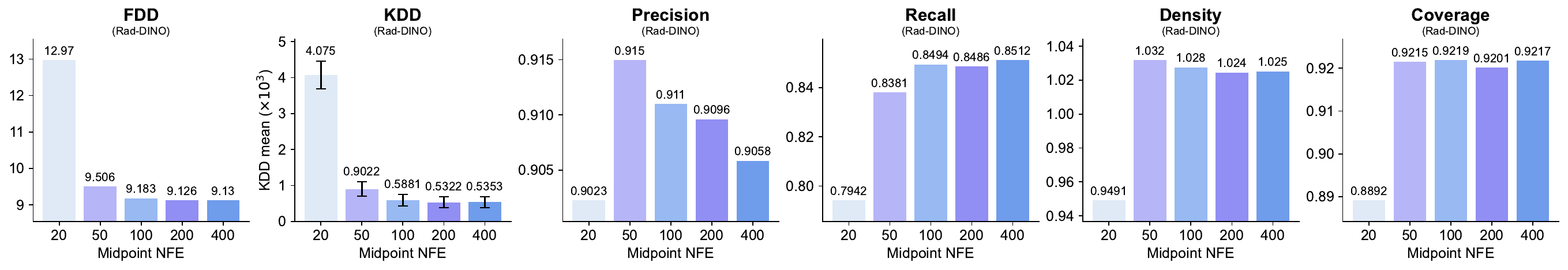}\\[1pt]
    \includegraphics[width=.999\textwidth]{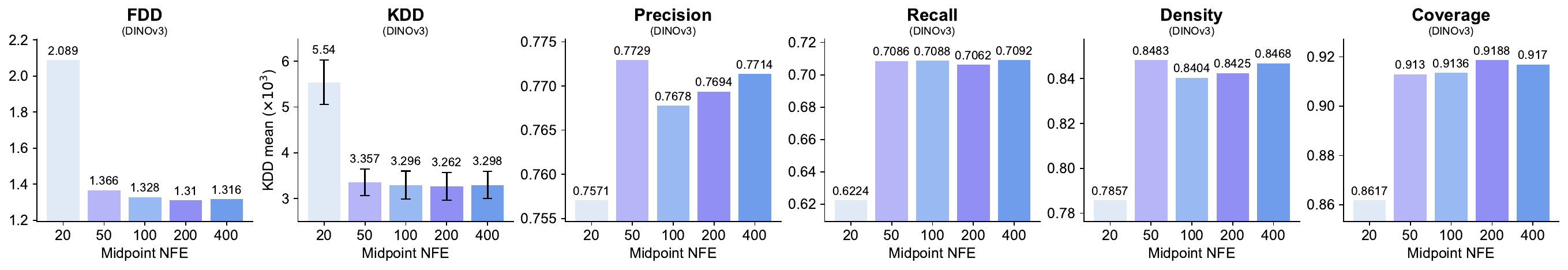}\\[1pt]
    \includegraphics[width=.999\textwidth]{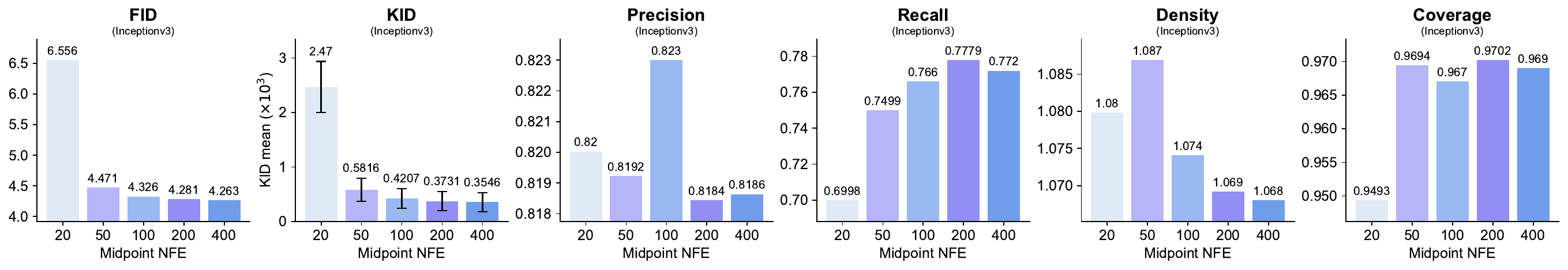}
    \caption{
    \textbf{Ablation of ODE solver \texttt{Midpoint} function evaluations (NFE)}. We used our RadiT XL model and the 5k-sample MIMIC-CXR split from the CXR7-1M test set as the reference. From top to bottom, each row shows results computed using Rad-DINO, DINOv3, and Inceptionv3 features. The \texttt{dopri5} ODE solver used to obtain the results in Table \ref{tab:sample_results} requires an average NFE of $\simeq$318.9.
    }
    \label{appfig:ode_solver_steps_ablation}
\end{figure}

\begin{table}[ht]
\centering
\small
\caption{
\textbf{Image editing performance for patient metadata.} Rad-DINO$^{\text{AP}}$ ROCAUC and MAE values for the observed (real) set of 5K images, and their edited versions. Values closer to the observed indicate better editing performance. 
}
\label{apptab:edit_auc_1}
\vspace{2pt}
\begin{tabular}{@{}llccccc@{}}
\toprule
 & & \multicolumn{4}{c}{\multirow{2}{*}{\textbf{ROCAUC}}} & {\textbf{MAE}} \\
 & & & & & & \footnotesize{(years)} \\
 & & Sex & Race & View & Dataset & Age \\
\midrule
\textbf{Observed} &  & 0.9986 & 0.9822 & 0.9980 & 1.0000 & 4.870 \\
\midrule
\textbf{Model} & \textbf{Edit} &  &  &  &  &  \\
\midrule
 RadiT XL & Sex & 0.9957 & 0.9822 & 0.9975 & 0.9966 & 6.581  \\
RadUNet$^{\textcolor{steelblue}{\text{pix}}}$ & Sex & 0.9886 & 0.9515 & 0.9969 & 0.9911 & 6.798 \\
\midrule
 RadiT XL & Race & 0.9984 & 0.8880 & 0.9973 & 0.9958 & 6.592 \\
RadUNet$^{\textcolor{steelblue}{\text{pix}}}$ & Race & 0.9977 & 0.7640 & 0.9969 & 0.9928 & 6.496 \\
\midrule
 RadiT XL & Age  & 0.9959 & 0.9636 & 0.9955 & 0.9927 & 7.492 \\
RadUNet$^{\textcolor{steelblue}{\text{pix}}}$ & Age & 0.9930 & 0.9450 & 0.9927 & 0.9837 & 9.267 \\
\midrule
 RadiT XL & View  & 0.9983 & 0.9849 & 0.9562 & 0.9580 & 6.399 \\
RadUNet$^{\textcolor{steelblue}{\text{pix}}}$ & View  & 0.9936 & 0.9360 & 0.9361 & 0.9232 & 7.171 \\
\midrule
 RadiT XL & Dataset  & 0.9971 & 0.9527 & 0.9957 & 0.9973 & 6.277 \\
RadUNet$^{\textcolor{steelblue}{\text{pix}}}$ & Dataset  & 0.9944 & 0.8653 & 0.9834 & 0.9423 & 7.096 \\
\midrule
 RadiT XL & Finding  & 0.9984 & 0.9816 & 0.9965 & 0.9962 & 6.487 \\
RadUNet$^{\textcolor{steelblue}{\text{pix}}}$ & Finding & 0.9977 & 0.9673 & 0.9965 & 0.9938 & 6.191 \\
\bottomrule
\end{tabular}
\end{table}

\begin{table}[ht]
\centering
\small
\caption{
\textbf{Image editing performance for clinical findings.} Rad-DINO$^{\text{AP}}$ ROCAUC for the observed (real) set of 5K images, and their edited versions. Values closer to the observed indicate better editing performance. n/a results are due to a lack of the negative class. AT = atelectasis, CA = cardiomegaly, CO = consolidation, ED = edema, EC = enlarged cardiomediastinum, FR = fracture, LL = lung lesion.
}
\label{apptab:edit_auc_2}
\vspace{2pt}
\begin{tabular}{@{}llccccccc@{}}
\toprule
 &  & \multicolumn{6}{c}{\textbf{ROCAUC}} \\
 &  & AT & CA & CO & ED & EC & FR & LL \\
\midrule
\textbf{Observed} &  & 0.8764 & 0.9587 & 0.9587 & 0.9618 & 0.929 & 0.9448 & 0.9438\\
\midrule
\textbf{Model} & \textbf{Edit} &  &  &  &  &  \\
\midrule
 RadiT XL & Sex & 0.7708 & 0.9537 & 0.9614 & 0.9383 & 0.9390 & 0.9170 & 0.8339 \\
RadUNet$^{\textcolor{steelblue}{\text{pix}}}$ & Sex & 0.7655 & 0.9430 & 0.9204 & 0.9548 & 0.9299 & 0.9097 & 0.8862 \\
\midrule
 RadiT XL & Race & 0.7887 & 0.9486 & 0.9369 & 0.9541 & 0.9328 & 0.9281 & 0.8928 \\
RadUNet$^{\textcolor{steelblue}{\text{pix}}}$ & Race & 0.8677 & 0.9336 & 0.9501 & 0.9581 & 0.9154 & 0.9194 & 0.9326 \\
\midrule
 RadiT XL & Age  & 0.8787 & 0.9464 & 0.9310 & 0.9571 & 0.9089 & 0.9149 & 0.8967 \\
RadUNet$^{\textcolor{steelblue}{\text{pix}}}$ & Age & 0.8402 & 0.9345 & 0.9282 & 0.9609 & 0.8962 & 0.9130 & 0.8921 \\
\midrule
 RadiT XL & View  & 0.7271 & 0.9397 & 0.9099 & 0.9104 & 0.9364 & 0.9077 & 0.8097 \\
RadUNet$^{\textcolor{steelblue}{\text{pix}}}$ & View  & 0.4827 & 0.8900 & 0.8762 & 0.8897 & 0.8823 & 0.8737 & 0.7229 \\
\midrule
 RadiT XL & Dataset  & n/a & 0.9735 & 0.9348 & 0.8645 & 0.9830 & 0.9810 & 0.6854 \\
RadUNet$^{\textcolor{steelblue}{\text{pix}}}$ & Dataset  & n/a & 0.9503 & 0.7795 & 0.8770 & 0.9278 & 0.9709 & 0.9213 \\
\midrule
 RadiT XL & Finding  & 0.6408 & 0.8942 & 0.7414 & 0.8338 & 0.7148 & 0.7436 & 0.6552 \\
RadUNet$^{\textcolor{steelblue}{\text{pix}}}$ & Finding & 0.6250 & 0.8946 & 0.7253 & 0.8237 & 0.6834 & 0.7046 & 0.6605 \\
\bottomrule
\end{tabular}
\end{table}

\begin{table}[ht]
\centering
\small
\caption{
\textbf{Image editing performance for clinical findings.} Rad-DINO$^{\text{AP}}$ ROCAUC values for the observed (real) set of 5K images, and their edited versions. Values closer to the observed indicate better editing performance. n/a results are due to a lack of the negative class. LO = lung opacity, NF = no finding, PE = pleural effusion, PO = pleural (other), PX = pneumothorax, SD = support devices.
}
\label{apptab:edit_auc_3}
\vspace{2pt}
\begin{tabular}{@{}llccccccc@{}}
\toprule
 &  & \multicolumn{6}{c}{\textbf{ROCAUC}} \\
 &  & LO & NF & PE & PO & PX & SD \\
\midrule
\textbf{Observed} &  & 0.8375 & 0.9903 & 0.9822 & 0.9275 & 0.9405 & 0.8888 \\
\midrule
\textbf{Model} & \textbf{Edit} &  &  &  &  &  \\
\midrule
 RadiT XL & Sex & 0.7993 & 0.9934 & 0.9766 & n/a & 0.9077 & 0.8255 \\
RadUNet$^{\textcolor{steelblue}{\text{pix}}}$ & Sex & 0.8148 & 0.9891 & 0.9673 &  & 0.9004 & 0.8362 \\
\midrule
 RadiT XL & Race & 0.7981 & 0.9886 & 0.9762 & n/a & 0.8961 & 0.8673 \\
RadUNet$^{\textcolor{steelblue}{\text{pix}}}$ & Race & 0.7913 & 0.9726 & 0.9719 &  & 0.9202 & 0.8394 \\
\midrule
 RadiT XL & Age  & 0.7997 & 0.9762 & 0.9714 & n/a & 0.8888 & 0.8627 \\
RadUNet$^{\textcolor{steelblue}{\text{pix}}}$ & Age & 0.7943 & 0.9676 & 0.9578 &  & 0.9033 & 0.8158 \\
\midrule
 RadiT XL & View  & 0.7219 & 0.9802 & 0.9697 & n/a & 0.8677 & 0.7380 \\
RadUNet$^{\textcolor{steelblue}{\text{pix}}}$ & View  & 0.7242 & 0.9596 & 0.9490 &  & 0.8161 & 0.7650 \\
\midrule
 RadiT XL & Dataset  & 0.8250 & 0.9286 & 0.9864 & n/a & 0.8568 & 0.9764 \\
RadUNet$^{\textcolor{steelblue}{\text{pix}}}$ & Dataset  & 0.5674 & 0.8948 & 0.9806 &  & 0.8495 & 0.8443 \\
\midrule
 RadiT XL & Finding  & 0.6006 & 0.9634 & 0.8927 & n/a & 0.8137 & 0.7640 \\
RadUNet$^{\textcolor{steelblue}{\text{pix}}}$ & Finding & 0.5943 & 0.9608 & 0.8760 &  & 0.7599 & 0.6640 \\
\bottomrule
\end{tabular}
\end{table}

\begin{table}[ht]
\centering
\small
\caption{
\textbf{Sample editing performance for patient metadata.} Rad-DINO$^{\text{AP}}$ ROCAUC and MAE values for the baseline set of 5K sampled images, and their edited versions. Values closer to the baseline indicate better editing performance. 
}
\label{apptab:sample_auc_1}
\vspace{2pt}
\begin{tabular}{@{}llccccc@{}}
\toprule
 & & \multicolumn{4}{c}{\multirow{2}{*}{\textbf{ROCAUC}}} & {\textbf{MAE}} \\
 & & & & & & \footnotesize{(years)} \\
 & & Sex & Race & View & Dataset & Age \\
\midrule
RadiT XL & Baseline & 0.9990 & 0.9734 & 0.9978 & 0.9957 & 6.424\\
RadUNet$^{\textcolor{steelblue}{\text{pix}}}$ & Baseline & 0.9968 & 0.9252 & 0.9954 &  0.9871 & 6.519 \\
\midrule
 RadiT XL & Sex &  0.9984 & 0.9765 & 0.9964 & 0.9961 & 6.425\\
RadUNet$^{\textcolor{steelblue}{\text{pix}}}$ & Sex &  0.9967 & 0.9237 & 0.9947 & 0.9868 & 6.514 \\
\midrule
 RadiT XL & Race &  0.9984 & 0.9212 & 0.9975 & 0.9951 & 6.568 \\
RadUNet$^{\textcolor{steelblue}{\text{pix}}}$ & Race &  0.9957 & 0.8450 & 0.9953 & 0.9834 & 6.669 \\
\midrule
 RadiT XL & Age  &  0.9952 & 0.9557 & 0.9955 & 0.9921 & 7.006 \\
RadUNet$^{\textcolor{steelblue}{\text{pix}}}$ & Age &  0.9914 & 0.9040 & 0.9926 & 0.9767 & 8.245 \\
\midrule
 RadiT XL & View  &  0.9981 & 0.9783 & 0.9641 & 0.9562 & 6.065 \\
RadUNet$^{\textcolor{steelblue}{\text{pix}}}$ & View  &  0.9959 & 0.9264 & 0.9662 & 0.9091 & 6.556 \\
\midrule
 RadiT XL & Dataset  &  0.9978 & 0.9555 & 0.9946 & 0.9973 & 6.198 \\
RadUNet$^{\textcolor{steelblue}{\text{pix}}}$ & Dataset  &  0.9957 & 0.9005 & 0.9921 & 0.9865 & 6.218 \\
\midrule
 RadiT XL & Finding  & 0.9986 & 0.9761 & 0.9977 & 0.9962 & 6.312 \\
RadUNet$^{\textcolor{steelblue}{\text{pix}}}$ & Finding & 0.9959 & 0.9159 & 0.9950 & 0.9857 & 6.587 \\
\bottomrule
\end{tabular}
\end{table}

\begin{table}[ht]
\centering
\small
\caption{
\textbf{Sample editing performance for patient metadata.} Rad-DINO$^{\text{AP}}$ ROCAUC values for the baseline set of 5K sampled images, and their edited versions. Values closer to the baseline indicate better editing performance.  n/a results are due to a lack of the negative class. AT = atelectasis, CA = cardiomegaly, CO = consolidation, ED = edema, EC = enlarged cardiomediastinum, FR = fracture, LL = lung lesion.
}
\label{apptab:sample_auc_2}
\vspace{2pt}
\begin{tabular}{@{}llccccccc@{}}
\toprule
 &  & \multicolumn{6}{c}{\textbf{ROCAUC}} \\
 &  & AT & CA & CO & ED & EC & FR & LL \\
 \midrule
\textbf{Model} & \textbf{Edit} &  &  &  &  &  \\
\midrule
RadiT XL & Baseline & 0.8101 & 0.9446 & 0.9225 & 0.9312 & 0.9268 & 0.9238 & 0.8992 \\
RadUNet$^{\textcolor{steelblue}{\text{pix}}}$ & Baseline & 0.7277 & 0.9343 & 0.9039 & 0.9284 & 0.9111 & 0.8814 & 0.7721 \\
\midrule
 RadiT XL & Sex &  0.7051 & 0.9452 & 0.9267 & 0.9208 & 0.9050 & 0.9230 & 0.8866 \\
RadUNet$^{\textcolor{steelblue}{\text{pix}}}$ & Sex &  0.5257 & 0.9225 & 0.9071 & 0.9245 & 0.9044 & 0.9040 & 0.8490 \\
\midrule
 RadiT XL & Race &  0.7230 & 0.9436 & 0.9352 & 0.9375 & 0.9133 & 0.9266 & 0.8384 \\
RadUNet$^{\textcolor{steelblue}{\text{pix}}}$ & Race &  0.6927 & 0.9200 & 0.9176 & 0.9323 & 0.9027 & 0.8785 & 0.7785 \\
\midrule
 RadiT XL & Age  &  0.7064 & 0.9303 & 0.9067 & 0.9144 & 0.9052 & 0.8787 & 0.8961 \\
RadUNet$^{\textcolor{steelblue}{\text{pix}}}$ & Age &  0.6420 & 0.9365 & 0.8940 & 0.9140 & 0.8710 & 0.8129 & 0.8436 \\
\midrule
 RadiT XL & View  &  0.5335 & 0.9426 & 0.9193 & 0.8644 & 0.8896 & 0.9183 & 0.8795 \\
RadUNet$^{\textcolor{steelblue}{\text{pix}}}$ & View  &  0.4173 & 0.9131 & 0.8810 & 0.8450 & 0.8453 & 0.8532 & 0.8012 \\
\midrule
 RadiT XL & Dataset  &  n/a & 0.9746 & 0.9414 & 0.8021 & 0.9833 & 0.9908 & 0.9663 \\
RadUNet$^{\textcolor{steelblue}{\text{pix}}}$ & Dataset  &  n/a & 0.9655 & 0.9069 & 0.8779 & 0.9388 & 0.9861 & 0.6461 \\
\midrule
 RadiT XL & Finding  & 0.6484 & 0.9002 & 0.7251 & 0.8033 & 0.7101 & 0.7459 & 0.6559 \\
RadUNet$^{\textcolor{steelblue}{\text{pix}}}$ & Finding & 0.6246 & 0.8821 & 0.6946 & 0.7784 & 0.7162 & 0.6673 & 0.6321 \\
\bottomrule
\end{tabular}
\end{table}

\begin{table}[ht]
\centering
\small
\caption{
\textbf{Sample editing performance for patient metadata.} Rad-DINO$^{\text{AP}}$ ROCAUC values for the baseline set of 5K sampled images, and their edited versions. Values closer to the baseline indicate better editing performance.  n/a results are due to a lack of the negative class. LO = lung opacity, NF = no finding, PE = pleural effusion, PO = pleural (other), PX = pneumothorax, SD = support devices.
}
\label{apptab:sample_auc_3}
\vspace{2pt}
\begin{tabular}{@{}llccccccc@{}}
\toprule
 &  & \multicolumn{6}{c}{\textbf{ROCAUC}} \\
 &  & LO & NF & PE & PO & PX & SD \\
 \midrule
\textbf{Model} & \textbf{Edit} &  &  &  &  &  \\
\midrule
RadiT XL & Baseline & 0.7802 & 0.9906 & 0.9571 & n/a & 0.8698 & 0.8598 \\
RadUNet$^{\textcolor{steelblue}{\text{pix}}}$ & Baseline & 0.7529 & 0.9858 & 0.9592 & n/a & 0.7547 & 0.6153\\
\midrule
 RadiT XL & Sex &  0.7790 & 0.9916 & 0.9669 &  & 0.8568 & 0.8117 \\
RadUNet$^{\textcolor{steelblue}{\text{pix}}}$ & Sex &  0.7231 & 0.9758 & 0.9530 & n/a & 0.7490 & 0.5620 \\
\midrule
 RadiT XL & Race &  0.8246 & 0.9925 & 0.9474 &  & 0.8890 & 0.8157 \\
RadUNet$^{\textcolor{steelblue}{\text{pix}}}$ & Race & 0.8121 & 0.9806 & 0.9597 & n/a & 0.7425 & 0.6163 \\
\midrule
 RadiT XL & Age  &  0.7401 & 0.9714 & 0.9530 &  & 0.8282 & 0.7773 \\
RadUNet$^{\textcolor{steelblue}{\text{pix}}}$ & Age &  0.7459 & 0.9661 & 0.9330 & n/a & 0.7420 & 0.5695 \\
\midrule
 RadiT XL & View  &  0.7718 & 0.9832 & 0.9599 &  & 0.8641 & 0.8459 \\
RadUNet$^{\textcolor{steelblue}{\text{pix}}}$ & View  &  0.7118 & 0.9577 & 0.9429 & n/a & 0.7321 & 0.6609 \\
\midrule
 RadiT XL & Dataset  &  0.8318 & 0.9316 & 0.9913 &  & 0.8831 & 0.2807 \\
RadUNet$^{\textcolor{steelblue}{\text{pix}}}$ & Dataset  &  0.7368 & 0.9020 & 0.9850 & n/a & 0.6404 & 0.5613 \\
\midrule
 RadiT XL & Finding  & 0.6215 & 0.9727 & 0.8866 &  & 0.7868 & 0.7750 \\
RadUNet$^{\textcolor{steelblue}{\text{pix}}}$ & Finding & 0.6220 & 0.9560 & 0.8830 & n/a & 0.6964 & 0.6265 \\
\bottomrule
\end{tabular}
\end{table}

\begin{figure*}[!t]
    \centering
    \includegraphics[width=\textwidth]{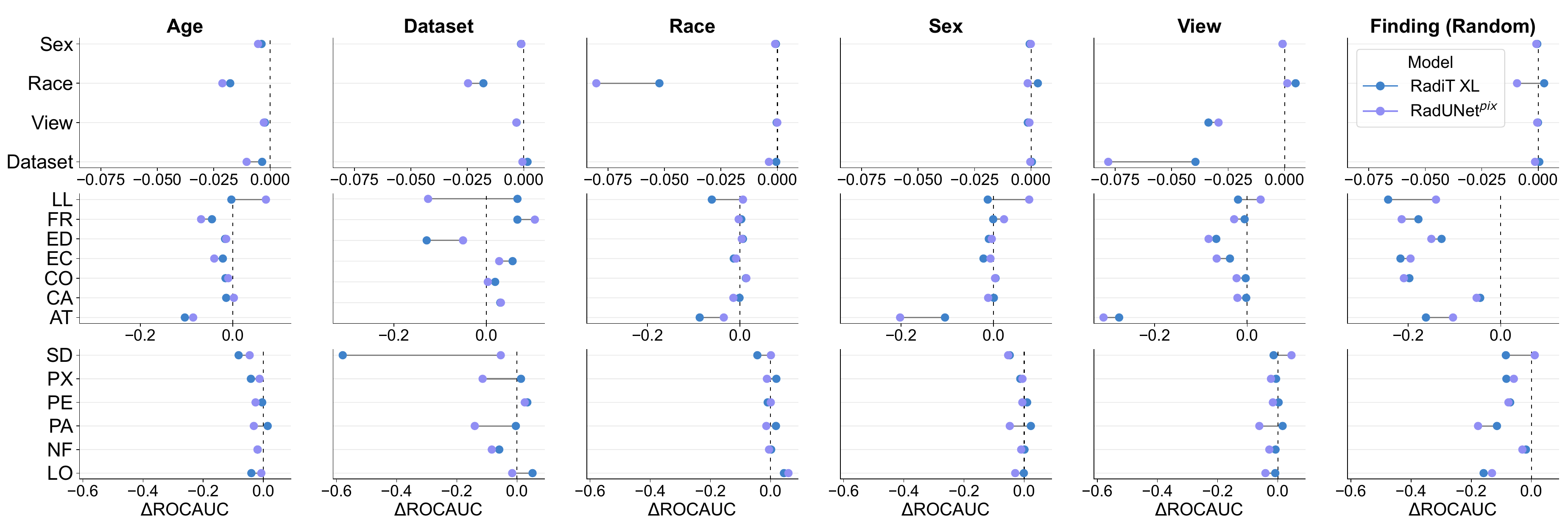}
    \caption{Difference in Rad-DINO$^{\text{AP}}$ ROCAUC between the 5K images sampled from our models, and their edited versions. Each column represents the attribute that was edited. $\Delta$ROCAUC values closer to zero indicate better editing performance. MAE values for Age are reported in Appendix \ref{appsec:experiments_control} (Table \ref{apptab:sample_auc_1}).}
    \label{appfig:sample_edit_auc}
\end{figure*}

\begin{figure*}[!ht]
    \centering
    \includegraphics[trim=0 9 0 5,clip,width=.98\textwidth]{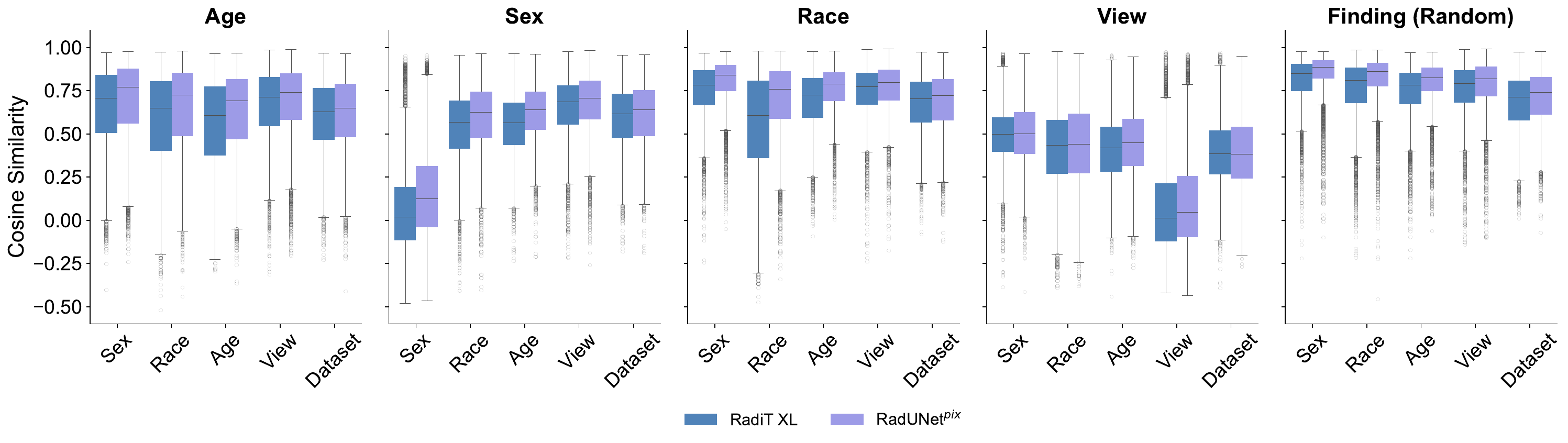}
    \caption{\textbf{Identity preservation comparison of our best latent- and pixel-space flow models.}
    Cosine similarity is measured between original and edited CXRs using Rad-DINO$^{\text{AP}}$ task conditioned embeddings. Titles show target edits; x-axis labels show task embeddings used. RadUNet$^{\textcolor{steelblue}{\text{pix}}}$ achieves higher cosine similarities, indicating better identity preservation but less effective editing.}
    \label{fig:cos_sim_bigmodels}
\end{figure*}

\begin{figure*}[!t]
    \centering
    \includegraphics[width=\textwidth]{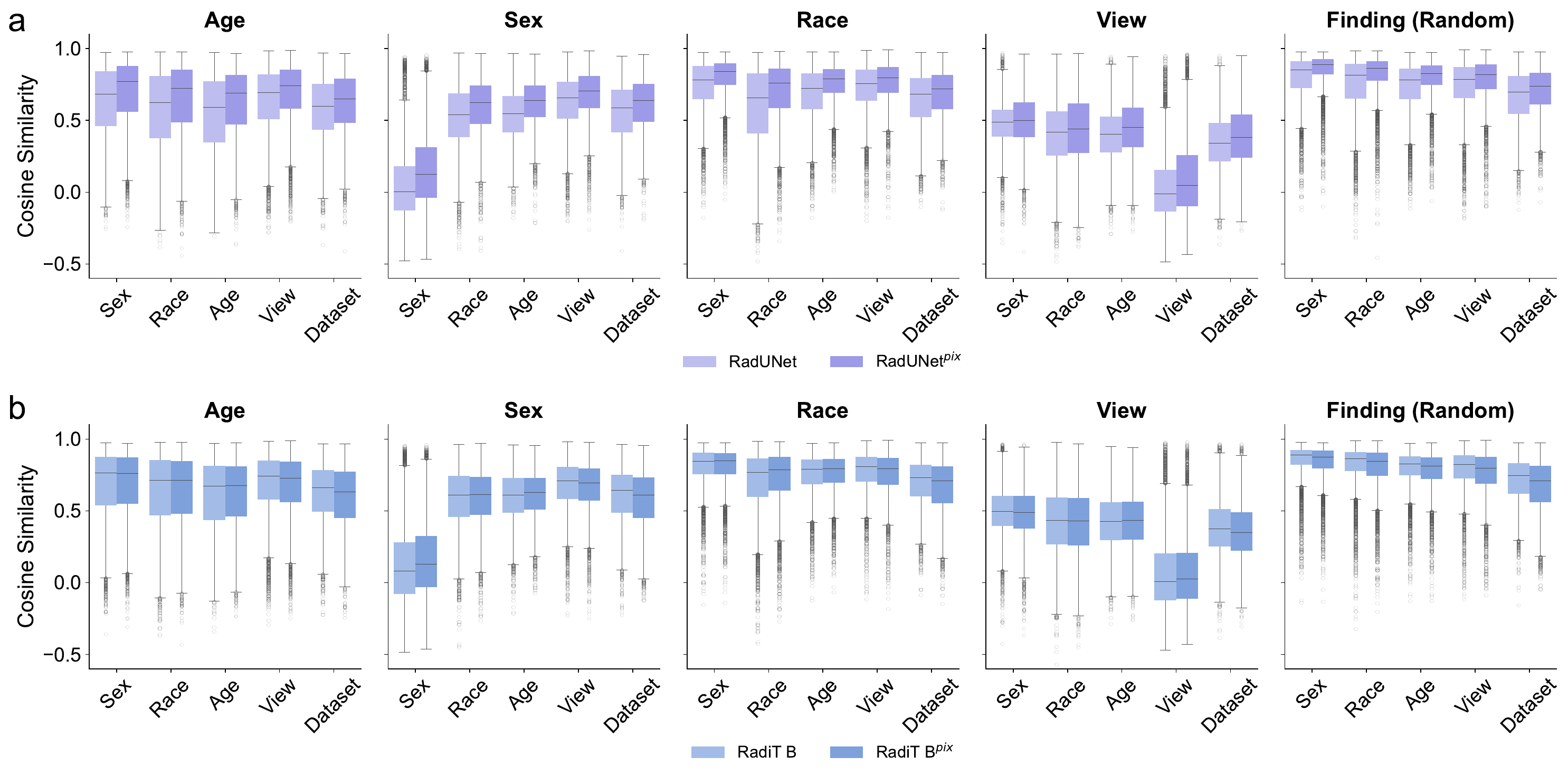}
    \caption{\textbf{Identity preservation comparison of latent- versus pixel-space flows with matched backbones.} Cosine similarity is measured between original and edited CXRs using Rad-DINO$^{\text{AP}}$ task conditioned embeddings. Titles show target edits; x-axis labels show task embeddings used. (\textbf{a}) UNet backbone, (\textbf{b}) Transformer backbone.}
    \label{appfig:cos_sim}
\end{figure*}

\clearpage
\subsection{Extra Qualitative Results}
\label{appsec:extra_quali}
\begin{figure}[!ht]
    \centering
    \includegraphics[width=\textwidth]{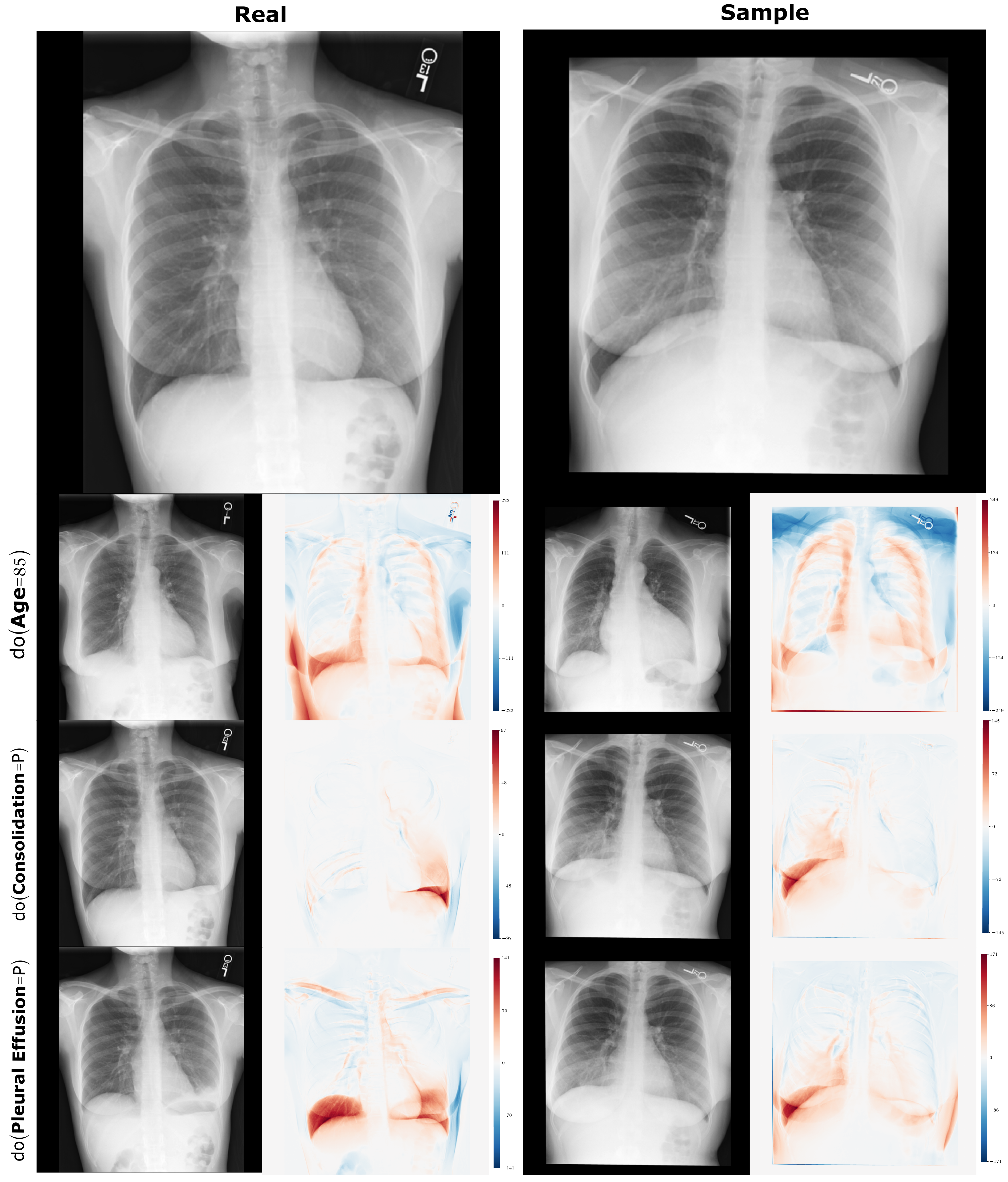}
    \caption{Edits performed on a real radiograph from CXR7-1M (left) and a synthetic radiograph sampled from RadiT XL (right). Both sets of images have the same metadata and clinical finding profile.
    }
    \label{appfig:edits_real_sample}
\end{figure}
\begin{figure}[!ht]
    \centering
    \includegraphics[width=.99\textwidth]{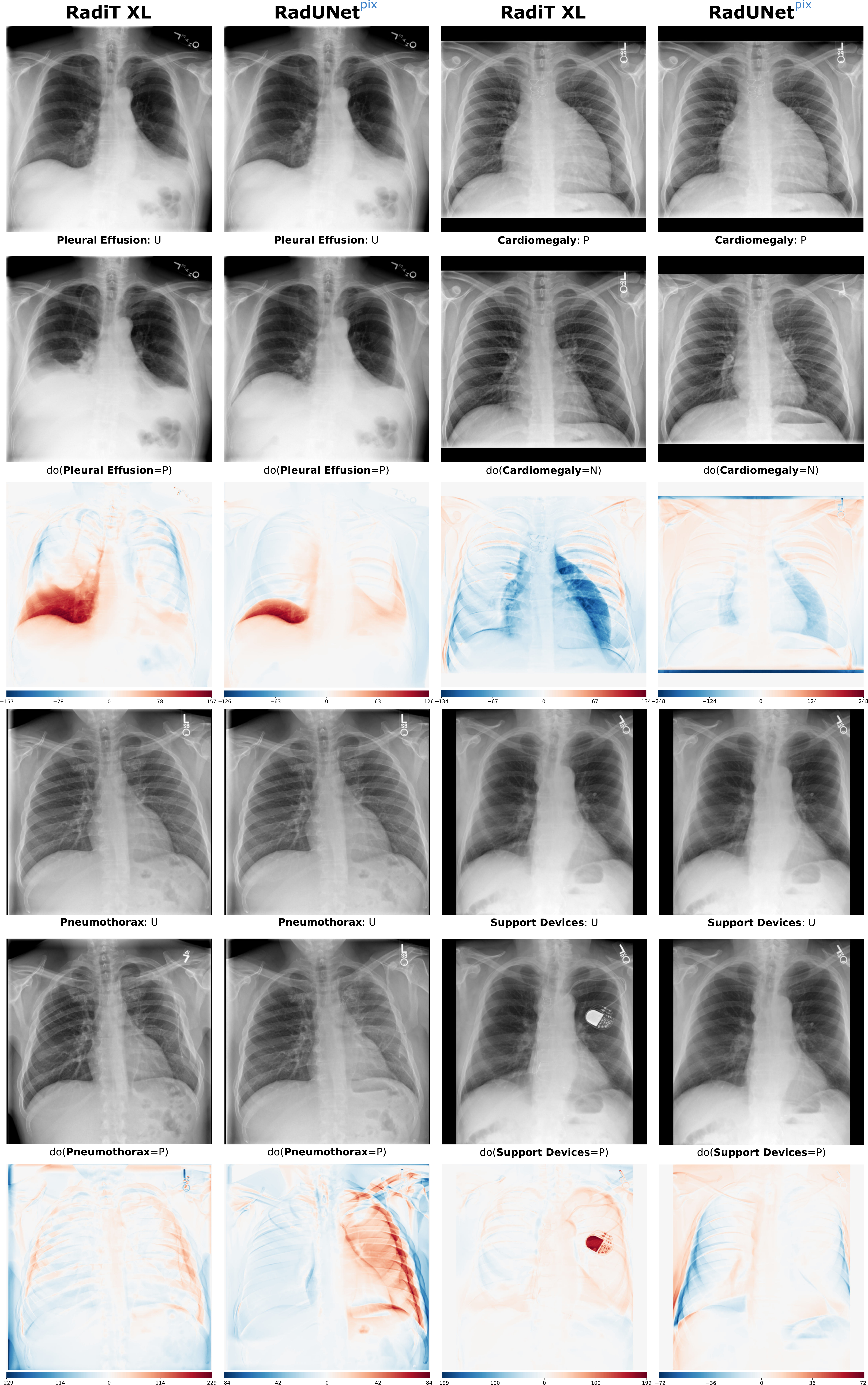}
    \caption{Comparing edit effectiveness of RadiT XL vs. RadUNet$^{\textcolor{steelblue}{\text{pix}}}$ on real radiographs.}
    \label{appfig:transformer_vs_unet_compressed}
\end{figure}
\begin{figure}[!ht]
    \centering
    \includegraphics[width=\textwidth]{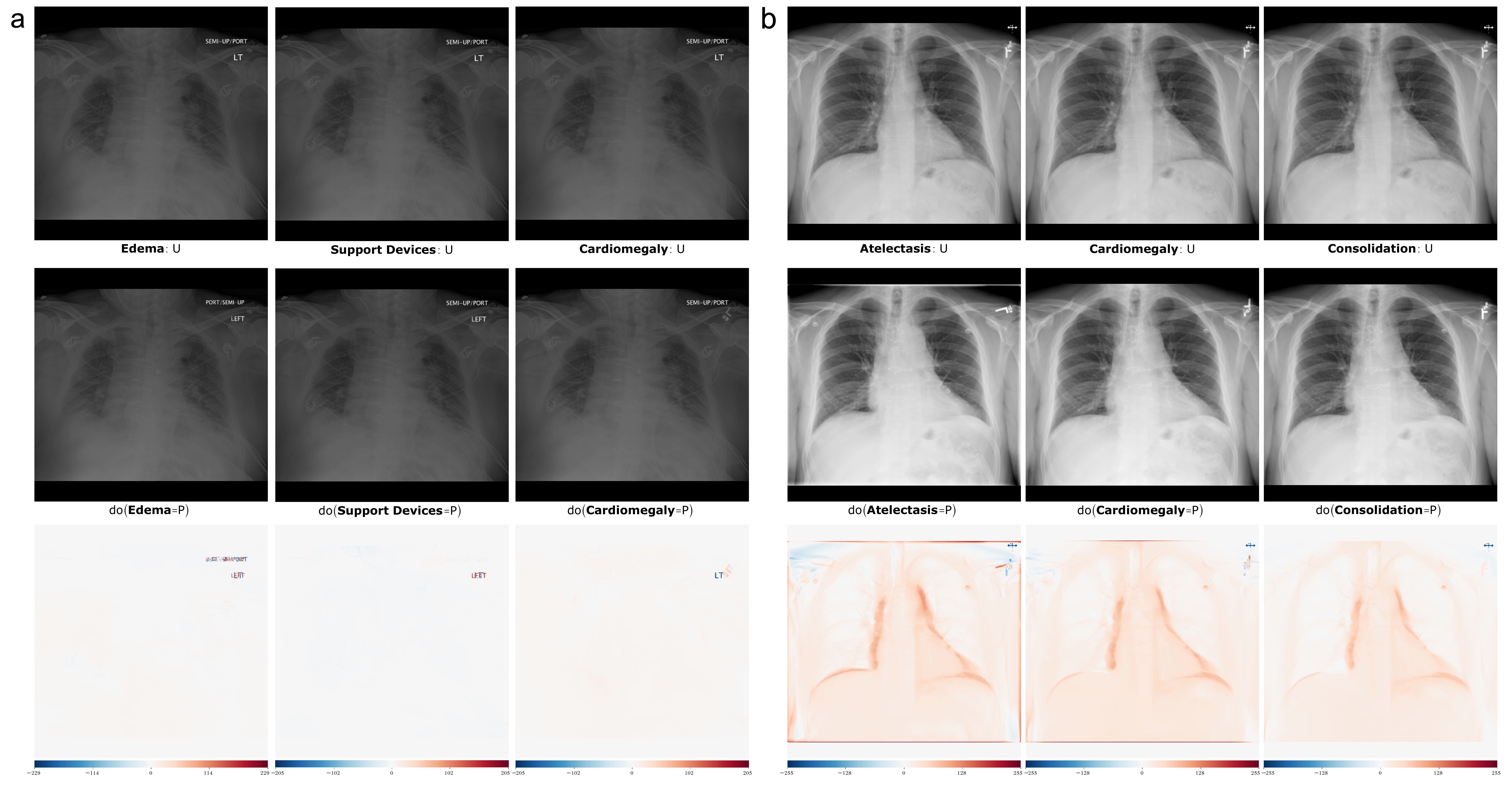}
    \caption{Examples of editing failure cases with RadiT XL. (\textbf{a}) No meaningful changes are made to the images. (\textbf{b}) The same changes are made for multiple clinical interventions.}
    \label{appfig:failure_cases}
\end{figure}
\begin{figure}[!ht]
    \centering
    \includegraphics[width=\textwidth]{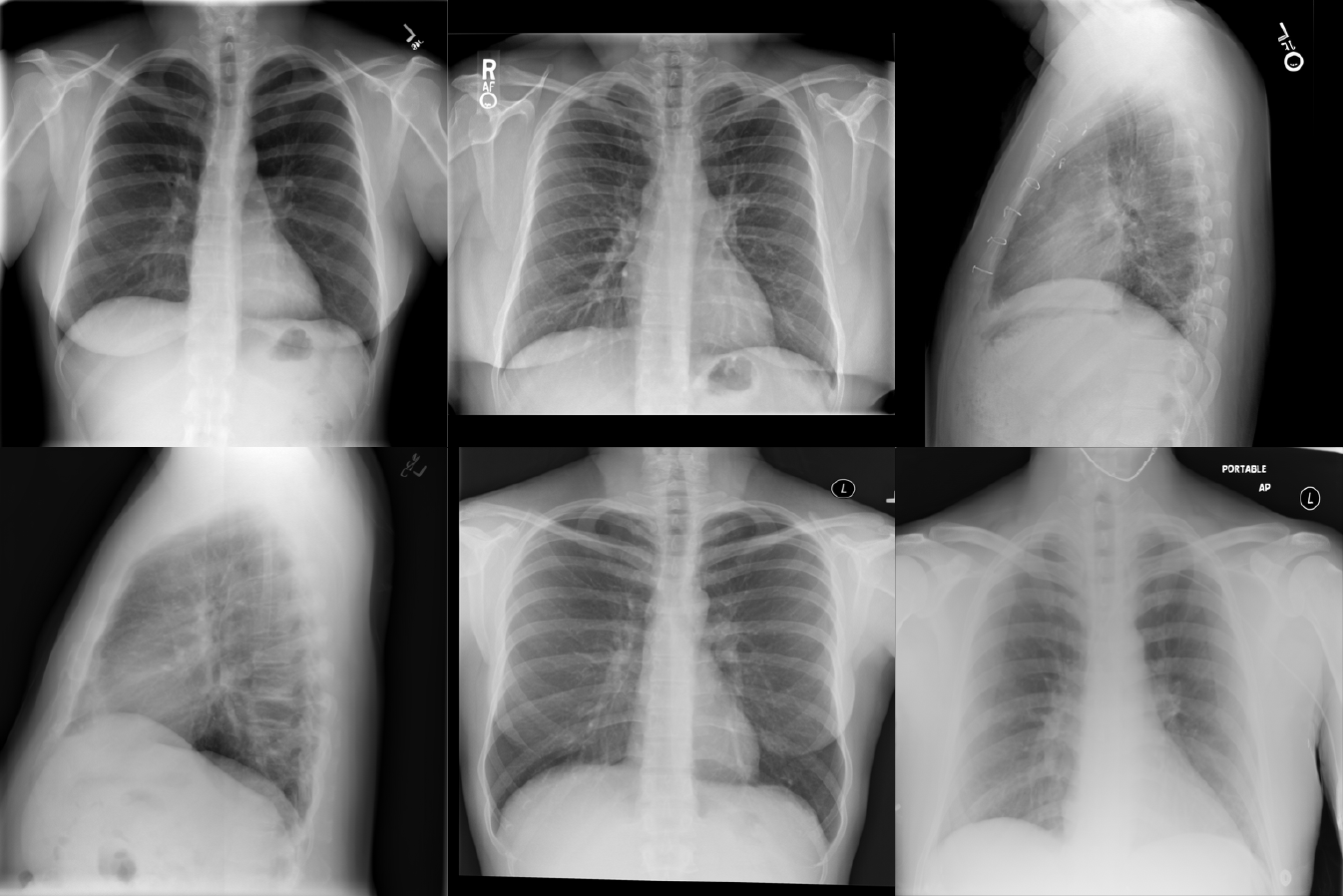}
    \caption{Synthetic 512${\times}$512 chest radiographs generated using our RadiT XL model.}
    \label{appfig:rand_samples_2}
\end{figure}
\begin{figure}[!ht]
    \centering
    \includegraphics[width=\textwidth]{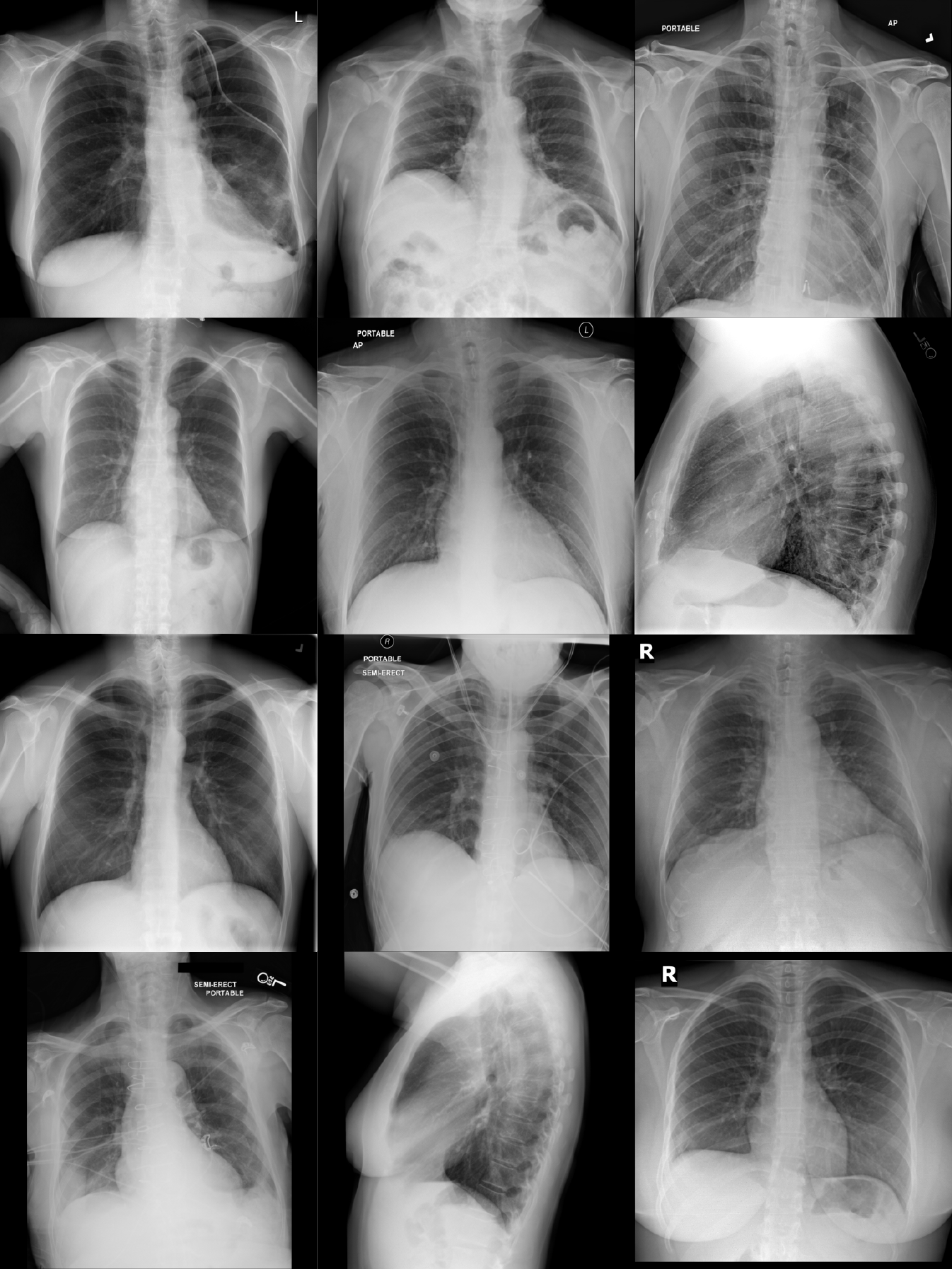}
    \caption{Synthetic 512${\times}$512 chest radiographs generated using our RadiT XL model.}
    \label{appfig:rand_samples_1}
\end{figure}
\begin{figure}
    \centering
    \includegraphics[width=\textwidth]{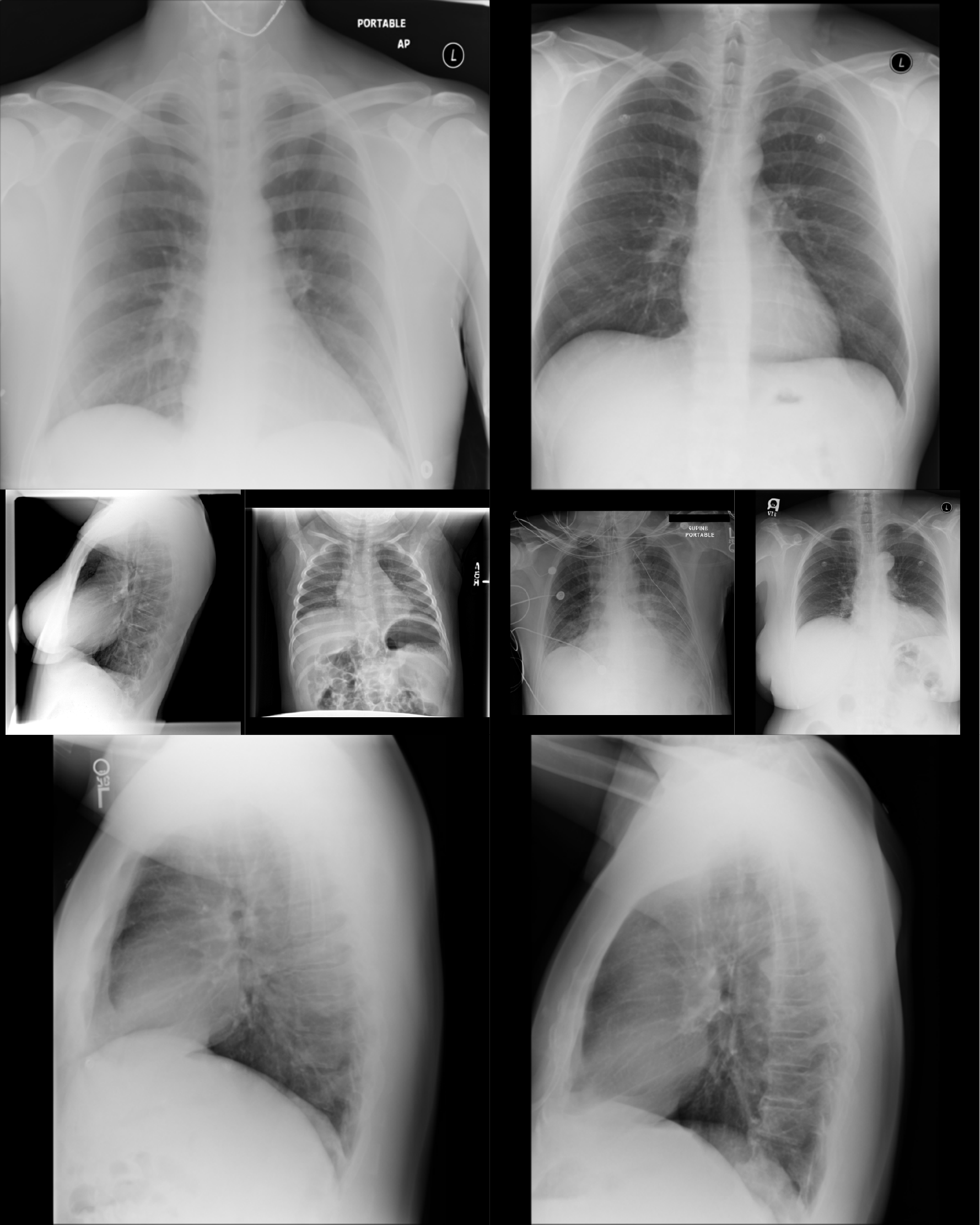}
    \caption{
    Synthetic 512${\times}$512 chest radiographs generated by our RadiT XL (1.3B) model, which all three clinical experts classified as real in a blinded real-vs-synthetic reader study.
    }
    \label{appfig:pred_reals}
\end{figure}


\end{document}